\documentclass{article}

\PassOptionsToPackage{numbers,compress}{natbib}

 \usepackage[preprint]{neurips_2026}

\usepackage[utf8]{inputenc} 
\usepackage[T1]{fontenc}    
\usepackage{hyperref}       
\usepackage{url}            
\usepackage{booktabs}       
\usepackage{amsfonts}       
\usepackage{nicefrac}       
\usepackage{microtype}      
\usepackage[table]{xcolor}
\usepackage{caption}
\usepackage{setspace}
\usepackage{subcaption}
\usepackage{amsmath}
\usepackage{amssymb}
\usepackage{algorithm}
\usepackage{algpseudocode}
\usepackage{multirow}
\usepackage{graphicx}
\usepackage{array}
\usepackage{tabularx}
\usepackage{wrapfig}

\definecolor{cornellred}{rgb}{0.7, 0.11, 0.11}
\definecolor{cadmiumgreen}{rgb}{0.0, 0.42, 0.24}
\definecolor{Blue9}{rgb}{0.098,0.3,0.9}

\hypersetup{
  linkcolor = cornellred,
  citecolor  = cadmiumgreen,
  colorlinks = true,
  urlcolor = Blue9
}

\makeatletter
\renewcommand{\@bottomtitlebar}{%
  \vskip 0.29in
  \vskip -\parskip
  \hrule height 1pt
  \vskip 0.38in
  \vskip -\parskip
}
\makeatother

\title{GuardMarkGS: Unified Ownership Tracing and Edit Deterrence for 3D Gaussian Splatting}

\author{%
\begin{tabular}{c}
\textbf{Utae Jeong}\textsuperscript{~1} \quad
\textbf{Jaewan Choi}\textsuperscript{~1} \quad
\textbf{Junseok Lee}\textsuperscript{~1} \\[2pt]
\textbf{Jongheon Jeong}\textsuperscript{~1} \quad
\textbf{Sang Ho Yoon}\textsuperscript{~2} \quad
\textbf{ByoungSoo Koh}\textsuperscript{~3} \quad
\textbf{Sangpil Kim}\textsuperscript{~1}\thanks{Corresponding Author} \\
\\[3pt]
{\normalfont \textsuperscript{1~}Korea University \quad
\textsuperscript{2~}KAIST \quad
\textsuperscript{3~}Hanshin University}
\end{tabular}%
}

\begin{document}

\renewcommand{\thefootnote}{\fnsymbol{footnote}}

\maketitle

\begin{abstract}
3D Gaussian Splatting (3DGS) is becoming a practical representation for novel view synthesis, but its growing adoption, together with rapid advances in instruction-driven 3DGS editing, also exposes a dual copyright risk: once a 3DGS-based asset is released, it can be used without permission and manipulated through 3D editing. Existing protection methods address only one side of this problem. Watermarking can trace ownership after unauthorized use, but it cannot prevent malicious editing. Adversarial edit-deterrence methods can disrupt editing, but they do not provide evidence of ownership. To the best of our knowledge, we present the first unified protection framework for 3DGS that jointly optimizes ownership tracing and unauthorized editing deterrence. Our framework combines a scene-wide watermarking objective over all Gaussians with an adversarial objective for edit deterrence. The adversarial branch combines latent-anchor separation, denoising-trajectory diversion, and cross-attention diversion to divert the editing trajectory, while an update-saliency-motivated Gaussian selection strategy assigns stronger adversarial updates to mask-selected Gaussians, improving the balance among watermark recovery, edit deterrence, and rendering fidelity. Experiments on scenes from Mip-NeRF 360 and Instruct-NeRF2NeRF demonstrate that the proposed framework achieves a favorable balance among bit accuracy, edit deterrence, and rendering quality. These results suggest that practical copyright protection of 3DGS-based assets can be more effectively addressed by integrating ownership tracing and unauthorized editing deterrence into a single optimization framework.
\end{abstract}

\section{Introduction}
\label{sec:intro}
\begin{figure*}[t]
    \centering
    \begin{minipage}{\textwidth}
        \centering
        \includegraphics[width=\linewidth]{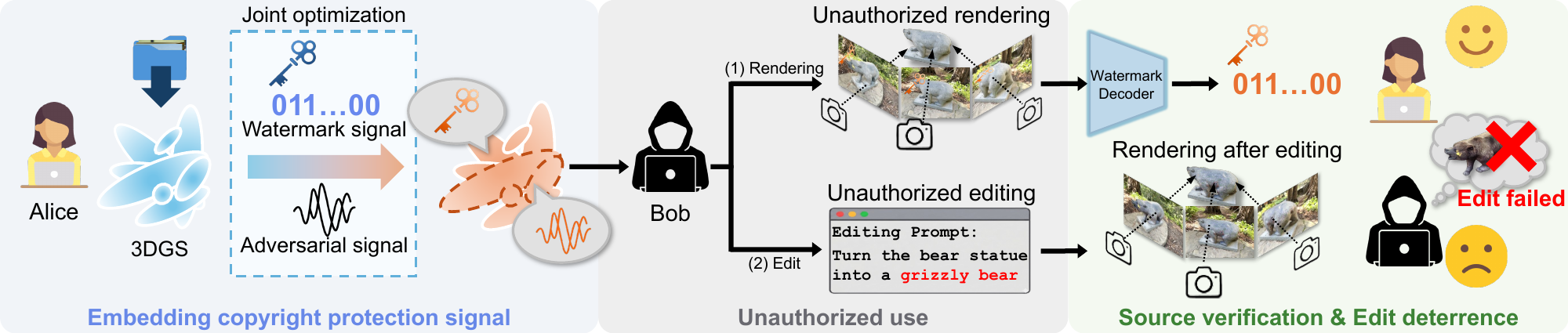}
        \caption{\textbf{Application scenario of GuardMarkGS.} We aim to identify ownership and prevent unauthorized editing by optimizing a 3DGS asset to embed watermark signals and adversarial noise. Left: Alice optimizes her 3DGS asset with watermark and adversarial signals. Middle: Bob performs unauthorized rendering and editing using Alice’s asset. Right: Alice verifies ownership by recovering the watermark with her decoder, while preventing Bob from obtaining the intended editing results.}
        \label{fig:teaser}
    \end{minipage}
\vspace{-15pt}    
\end{figure*}

3D Gaussian Splatting (3DGS)~\citep{3dgs} has emerged as a powerful representation for novel view synthesis~\citep{nerf,instantngp,scaffoldgs}, combining photorealistic rendering quality with real-time rendering speed. As 3DGS assets are increasingly used in AR/VR, digital content production, and other commercial pipelines, copyright protection becomes a central concern. In practice, a released 3DGS scene can be redistributed without authorization and manipulated by 3D editing pipelines~\citep{nerf2nerf,dge,gaussianeditor,gaussctrl,editsplat}. Recent advances in text-driven editing~\citep{clipnerf,dreameditor,voxe} further make it possible to produce visually convincing derivatives from released assets using only text prompts. As illustrated in Fig.~\ref{fig:teaser}, this creates a practical threat scenario in which an owner may lose control over both proving ownership and preserving the integrity of the content.

These observations indicate that copyright protection for 3DGS should not be formulated only as watermark insertion after release. A watermark can support ownership tracing after unauthorized use, but it cannot prevent the released asset from being transformed into a semantically altered derivative. A practical protection mechanism should therefore preserve evidence of ownership while also deterring unauthorized instruction-driven editing.

Existing protection strategies are split into two largely disconnected families. Passive protection methods, such as ownership tracing via digital watermarking~\citep{hidden,stablesignature,treering,robin,steganerf,copyrnerf,waterf}, identify the source of an asset after unauthorized use or redistribution, but their role is fundamentally post-hoc. By contrast, active protection methods based on adversarial optimization~\citep{photoguard,advdm,advpaint,mist,glaze,degauss} aim to make unauthorized editing difficult, but they do not embed a recoverable ownership signal for later verification. Thus, watermark-only methods focus on traceability, whereas adversarial edit-deterrence methods focus on deterrence, and neither alone fully addresses the practical copyright threat in Fig.~\ref{fig:teaser}.

However, unifying these two protection goals for 3DGS remains largely unexplored and is particularly challenging in the 3D setting. Reliable watermark recovery usually benefits from scene-wide optimization, because the ownership signal should remain detectable from many rendered viewpoints~\citep{3dgsw,gaussianmarker,guardsplt}. In contrast, imposing adversarial signals uniformly on the same scene-wide Gaussian support can unnecessarily perturb regions that are important for watermark recovery and rendering quality. This issue is not resolved by a straightforward composition of existing objectives: as shown in our experiments, directly adding DEGauss-style adversarial terms~\citep{degauss} to a 3DGSW-style watermarking objective~\citep{3dgsw} substantially weakens watermark recovery and rendering fidelity while providing limited edit deterrence. Thus, successful joint optimization must account for the different spatial and functional characteristics of watermarking and adversarial edit deterrence.

To address these issues, we propose \textbf{GuardMarkGS}, a unified protection framework that jointly optimizes ownership tracing and unauthorized editing deterrence within a single protected 3DGS. GuardMarkGS assigns distinct roles to the two protection signals within a single coordinated optimization. The watermarking branch preserves watermark signal through a scene-wide objective over rendered views, while the adversarial branch assigns stronger edit-deterrence updates to mask-selected Gaussians identified by update-saliency-motivated selection. We further introduce parameter-role-aware adversarial modulation, which scales adversarial gradients according to the rendering sensitivity of each Gaussian parameter. This selective and role-aware allocation reduces unnecessary interference with watermark recovery and rendering fidelity, which is the key difficulty in naive joint optimization. To further disrupt unauthorized editing, the adversarial branch combines latent-anchor separation with denoising-trajectory and cross-attention diversion terms~\citep{ldm}, targeting the latent representation, denoising direction, and prompt-image alignment used by editors. This design provides a unified mechanism for balancing watermark recovery, edit deterrence, and rendering fidelity. Under sUCPS, a unified score that jointly evaluates traceability, edit deterrence, and fidelity, GuardMarkGS improves over the best existing baseline by 10.7\%. Our main contributions are as follows:
\begin{itemize}
    \item We propose a unified dual-branch framework that jointly optimizes scene-wide ownership tracing and unauthorized edit deterrence for a single protected 3DGS representation.
    \item We introduce update-saliency-motivated Gaussian selection and parameter-role-aware adversarial modulation, which allocate stronger adversarial updates to defense-effective Gaussians while reducing unnecessary degradation to watermark recovery and rendering fidelity.
    \item We introduce an adversarial edit-deterrence objective that integrates latent-anchor separation with denoising-trajectory and cross-attention diversion, and experimentally show that the proposed method achieves a sound balance among traceability, edit deterrence, and fidelity.
\end{itemize}

\section{Related works}
\label{sec:relatedwork}
\textbf{3D Gaussian Splatting representation.} 3D Gaussian Splatting (3DGS)~\citep{3dgs} has become a widely used representation for novel view synthesis by modeling a scene with explicit anisotropic Gaussian primitives and rendering them efficiently through differentiable splatting. Its combination of photorealistic quality and real-time rendering makes 3DGS increasingly practical for interactive 3D applications, digital content production, and downstream asset reuse~\citep{mipsplatting,sugar,twodgs,gof,dreamgaussian,gaussiandreamer}. At the same time, the explicit Gaussian parameters make 3DGS directly editable and easy to redistribute. Once a 3DGS asset is released, its parameters can be copied, modified, or fine-tuned by downstream users, which creates a need for copyright protection mechanisms that preserve rendering fidelity while enabling ownership tracing of the released representation or protecting it against unauthorized manipulation.

\textbf{Copyright protection for 3D Gaussian Splatting.} Recent 3DGS copyright protection methods mostly follow a passive paradigm. GaussianMarker~\citep{gaussianmarker}, GuardSplat~\citep{guardsplt}, and 3DGSW~\citep{3dgsw} embed watermark signals into Gaussian parameters while attempting to preserve rendering quality. These methods support post-hoc ownership tracing: after an asset is leaked or redistributed, the owner can render images from the protected scene and decode a message to verify provenance. However, they do not explicitly prevent a malicious user from applying instruction-driven 3D edits~\citep{nerf2nerf,dge,gaussctrl,editsplat} to the released asset. As a result, they solve the tracing problem but not the deterrence problem: ownership can be verified after misuse, but the released content may already have been semantically altered.

\textbf{Defenses against instruction-driven editing.} A separate line of research studies active defenses~\citep{photoguard,advdm,advpaint,mist,glaze,antidreambooth} against diffusion-based image editing~\citep{ldm,instructpix2pix,ddpm,cfg}. These methods inject adversarial perturbations that disrupt latent diffusion guidance or denoising trajectories, thereby lowering edit success. Extending this idea to 3D assets is challenging because 3D editing pipelines repeatedly render the current scene from multiple viewpoints, edit the rendered images, and optimize the 3D representation to match the edited views. Perturbations that are effective for a single image may not remain consistent across viewpoints or may be weakened during iterative 3D optimization. DEGauss~\citep{degauss} extends adversarial protection to 3DGS and is, to our knowledge, the most relevant defense-only baseline in the 3D setting. However, DEGauss does not provide ownership tracing, and its whole-scene adversarial protection can noticeably degrade rendering quality. 

\textbf{3D Gaussian Splatting editing pipelines.} Text-driven 3DGS editing has advanced rapidly through pipelines such as DGE~\citep{dge} and GaussianEditor~\citep{gaussianeditor}. These methods repeatedly render the current 3D representation, edit the views with an image editing model~\citep{instructpix2pix}, and fit the 3DGS back to the edited results. This iterative render-edit-update cycle makes 3DGS editing powerful, but it also exposes a practical pathway for unauthorized manipulation. If an unauthorized user can access the released 3DGS model, they can produce high-quality malicious variants with only a text prompt. Thus, the same properties that make 3DGS useful for legitimate editing also make it vulnerable to misuse.

The above studies address important but incomplete aspects of 3DGS protection. Watermarking methods provide ownership evidence, but they do not actively deter unauthorized edits. Adversarial defense methods can suppress instruction-driven editing, but they do not provide recoverable ownership signals. Our work targets this missing intersection. Rather than treating ownership tracing and edit deterrence as separate goals, we formulate them as a unified optimization problem for a single protected 3DGS model. A scene-wide watermarking objective preserves ownership evidence across views, while an update-saliency-motivated adversarial objective allocates stronger updates to mask-selected Gaussian regions. This design provides a practical protection framework that supports ownership tracing, resists unauthorized editing, and simultaneously accounts for rendering quality.

\section{Problem setup}
\label{sec:problemsetup}
Let $\mathcal{G} = \{g_i\}_{i=1}^{N}$ denote a 3DGS scene composed of $N$ Gaussians, where each Gaussian contains geometric and appearance parameters such as position, scale, rotation, opacity, and view-dependent features. Rendering the scene from a camera viewpoint $v$ yields an image $I_v = \mathcal{R}_v(\mathcal{G})$, where $\mathcal{R}_v(\cdot)$ denotes the differentiable 3DGS renderer for viewpoint $v$.

Our goal is to optimize a protected scene $\mathcal{G}^{\mathrm{prot}}$ that satisfies three requirements simultaneously. First, it must carry an ownership message $\mathbf{m} \in \{0,1\}^{k}$, where $k$ denotes the message length, that can be decoded from rendered views with high bit accuracy. Second, under the threat scenario illustrated in Fig.~\ref{fig:teaser}, in which the released asset is subjected to editing with an edit instruction $\psi$, the edited outputs should deviate from the ideal editing results. Third, it must preserve the visual quality of the original 3DGS scene in novel-view rendering. Formally, the desired protected scene should satisfy
\begin{align}
    \text{Tracing:} &\quad \mathcal{D}_{\mathrm{msg}}\big(\mathcal{R}_v(\mathcal{G}^{\mathrm{prot}})\big) \rightarrow \mathbf{m}, \\
    \text{Deterrence:} &\quad \mathcal{E}\big(\mathcal{G}^{\mathrm{prot}}, \psi\big) \not\approx \mathcal{E}(\mathcal{G}, \psi), \\
    \text{Fidelity:} &\quad \mathcal{R}_v(\mathcal{G}^{\mathrm{prot}}) \approx \mathcal{R}_v(\mathcal{G}),
\end{align}
where $\mathcal{D}_{\mathrm{msg}}$ is the watermark decoder, $\mathcal{E}$ denotes the instruction-driven editing pipeline considered in our threat scenario, and $\psi$ denotes the edit instruction.

The central difficulty is to jointly satisfy these requirements under the fixed scene-wide support needed for watermark. Uniformly applying adversarial perturbations over the same support can interfere with watermark recovery, edit deterrence, and rendering fidelity. Moreover, unlike image-domain attacks with direct pixel-space budgets~\citep{photoguard,advdm,mist,glaze}, 3DGS perturbations are applied to Gaussian parameters with different rendering roles, while fidelity is observed only after rendering. Thus, our framework must control both the spatial support and parameter-role sensitivity of adversarial updates.

\section{Method}
\label{sec:method}
\subsection{Overview}
\label{subsec:overview}
\begin{figure*}[t]
    \centering
    \begin{minipage}{\textwidth}
        \centering
        \includegraphics[width=\linewidth]{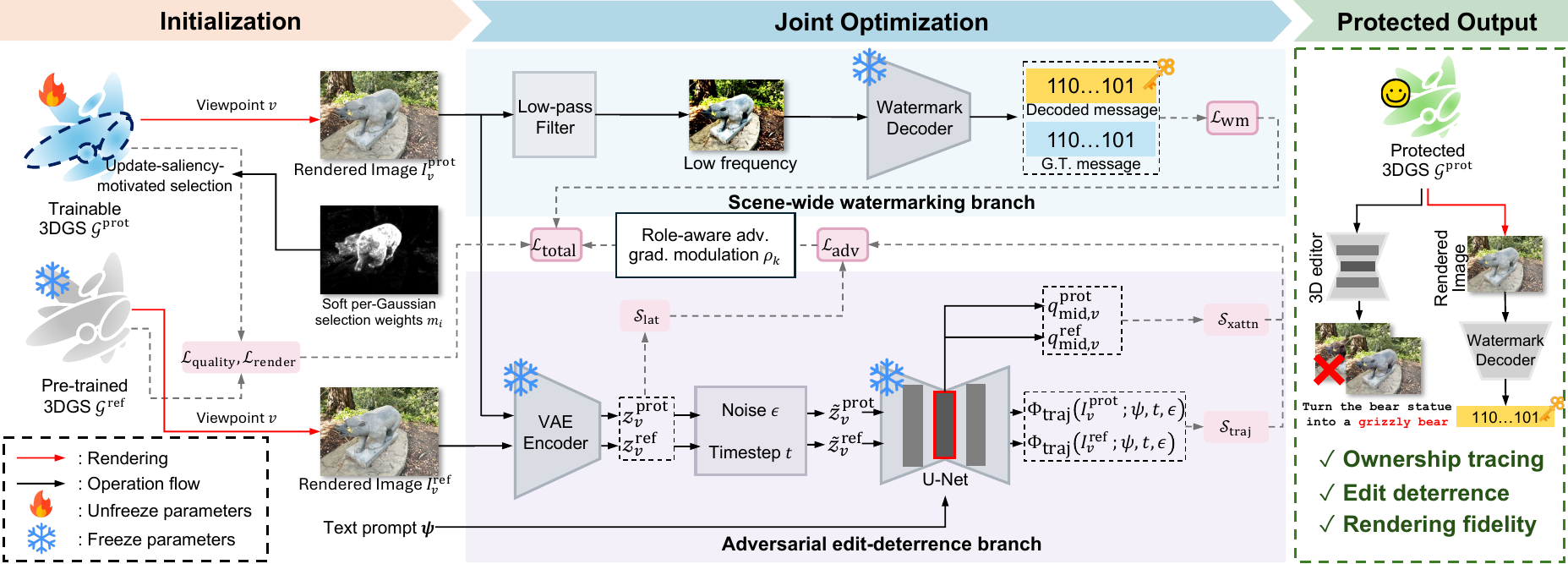}
        \caption{\textbf{Overview of GuardMarkGS.} Left: We initialize 3DGS from a pretrained scene, keep a reference, and compute soft per-Gaussian weights via update-saliency-motivated Gaussian selection. Middle: The watermarking branch embeds watermark signals, while the adversarial branch uses latent-anchor separation, denoising-trajectory diversion, and cross-attention diversion. These terms use rendering constraints and softly weighted adversarial gradients modulated by parameter role. Right: The optimized 3DGS preserves fidelity, supports ownership tracing, and deters editing.}
        \label{fig:overview}
    \end{minipage}
\vspace{-15pt}
\end{figure*}
Fig.~\ref{fig:overview} illustrates the proposed framework. GuardMarkGS consists of two optimization branches: a scene-wide watermarking branch for ownership tracing, described in Sec.~\ref{subsec:watermark}, and an adversarial edit-deterrence branch, described in Sec.~\ref{subsec:gaussian_selection}. At iteration $t$, we sample a training view set $\mathcal{V}_t$, which is shared by both branches. The watermarking branch optimizes all rendered views in $\mathcal{V}_t$ to embed a recoverable ownership message into the protected 3DGS scene. For the adversarial branch, we precompute a fixed soft Gaussian mask from text-guided multi-view segmentation before optimization. This mask controls the adversarial update strength of each Gaussian, assigning stronger edit-deterrence updates to mask-selected Gaussians while allowing surrounding Gaussians to receive weaker updates. We further modulate adversarial gradients according to the rendering role of each Gaussian parameter, mitigating the mismatch between Gaussian-parameter perturbations and rendered-image fidelity. We optimize the protected scene with the following overall objective:
\begin{equation}
    \mathcal{L}_{\mathrm{total}} = \mathcal{L}_{\mathrm{wm}} + \lambda_{\mathrm{adv}}\,\mathcal{L}_{\mathrm{adv}}.
    \label{eq:total}
\end{equation}
where $\mathcal{L}_{\mathrm{wm}}$ denotes the scene-wide watermarking objective for ownership tracing, $\mathcal{L}_{\mathrm{adv}}$ denotes the adversarial objective for edit deterrence, and $\lambda_{\mathrm{adv}}$ controls the relative strength of the edit-deterrence branch. The watermarking objective is applied scene-wide, whereas adversarial gradients are scaled by soft mask weights and parameter-role coefficients, enabling stronger updates on mask-selected Gaussians and more conservative perturbations for visually sensitive parameter groups.
\begin{figure*}[t]
    \centering
    \begin{minipage}{\textwidth}
        \centering
        \begin{subfigure}[t]{0.485\linewidth}
            \centering
            \makebox[\linewidth][c]{%
                \includegraphics[height=0.11\textheight]{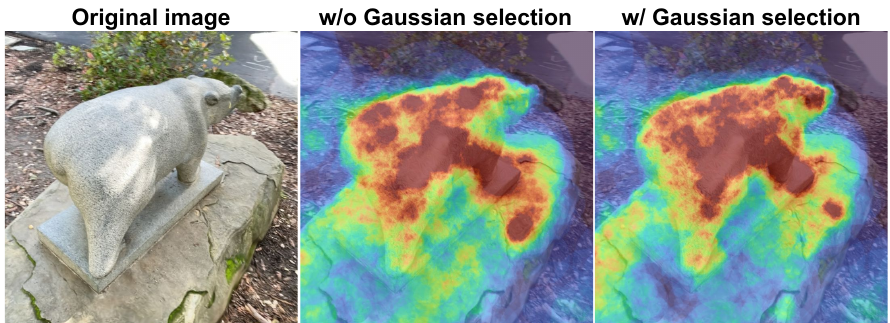}
            }
            \caption{\textit{bear}}
            \label{fig:gaussian_update_bear}
        \end{subfigure}
        \hspace{0.01\linewidth}
        \begin{subfigure}[t]{0.485\linewidth}
            \centering
            \makebox[\linewidth][c]{%
                \includegraphics[height=0.11\textheight]{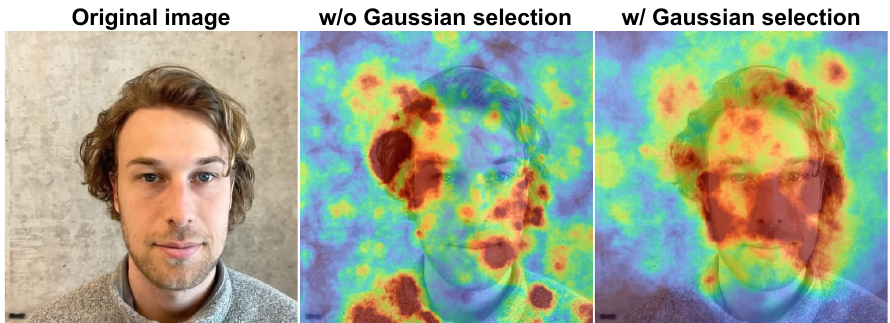}
            }
            \caption{\textit{face}}
            \label{fig:gaussian_update_face}
        \end{subfigure}
        \vspace{-5pt}
        \caption{\textbf{Visualization of adversarially updated Gaussians.} Without the proposed Gaussian selection, the adversarial objective naturally forms regions of large adversarial updates around the main scene objects in easier-to-defend scenes such as \textit{bear}. However, in more difficult-to-defend scenes such as \textit{face}, these update patterns become spatially scattered. Based on this observation, we use update-saliency-motivated selection to select Gaussians where defense-effective updates are concentrated and assign them stronger adversarial updates.}
        \label{fig:gaussian_update_visualization}
    \end{minipage}
\vspace{-15pt}
\end{figure*}

\subsection{Scene-wide watermark optimization}
\label{subsec:watermark}
The watermark branch follows an iterative message recovery process over rendered protected views using a pre-trained watermark decoder~\citep{hidden}. Since low-frequency components are generally more robust to common distortions, we decode the watermark from the low-frequency wavelet band of each rendered view~\citep{wm_lowfreq}. We define the watermark objective as follows:
\begin{equation}
    \mathcal{L}_{\mathrm{wm}} = \lambda_{\mathrm{msg}}\,\mathcal{L}_{\mathrm{msg}} + 
    \lambda_{\mathrm{quality}}\,\mathcal{L}_{\mathrm{quality}},
    \label{eq:lwm}
\end{equation}
where $\lambda_{\mathrm{msg}}$ and $\lambda_{\mathrm{quality}}$ control the relative weights of message recovery and rendering fidelity. Both $\mathcal{L}_{\mathrm{msg}}$ and $\mathcal{L}_{\mathrm{quality}}$ are computed for each rendered view in $\mathcal{V}_t$. Here, $\mathcal{L}_{\mathrm{msg}}$ is binary cross entropy between decoded bits and the target message, and $\mathcal{L}_{\mathrm{quality}}$ combines an $\ell_1$ term and an LPIPS~\citep{lpips} term between the protected renders and the corresponding frozen reference renders. We optimize this watermarking objective scene-wide because the watermark signal should remain recoverable from diverse viewpoints and robust to realistic image- or model-level distortions.

\subsection{Adversarial edit deterrence}
\label{subsec:gaussian_selection}
\subsubsection{Update-saliency-motivated Gaussian selection}
\label{subsec:update-saliency-selection}
Fig.~\ref{fig:gaussian_update_visualization} visualizes the Gaussians whose parameters are strongly updated by the adversarial terms later defined in Eq.~\eqref{eq:adv}. Without the proposed Gaussian selection, easier-to-defend scenes such as \textit{bear} naturally show concentrated update-salient patterns around the main scene objects, whereas more difficult-to-defend scenes such as \textit{face} exhibit irregular and spatially scattered updates. This observation suggests that effective edit deterrence benefits from concentrating the adversarial budget on a selected Gaussian subset rather than perturbing the entire scene uniformly. Motivated by this observation, we precompute text-guided 2D semantic masks~\citep{sam} before optimization and convert them into per-Gaussian mask scores by accumulating the mask values weighted by each Gaussian's rasterized pixel contribution. For each Gaussian $g_i$, we compute
\begin{equation}
    s_i = \frac{\sum_{v \in \mathcal{V}_{\mathrm{mask}}}\sum_{p} a_{i,v}(p)\,M_v(p)}{\sum_{v \in \mathcal{V}_{\mathrm{mask}}}\sum_{p} a_{i,v}(p) + \epsilon},
    \label{eq:score}
\end{equation}
where $M_v$ is the 2D mask at view $v$ and $a_{i,v}(p)$ denotes the rasterized contribution of Gaussian $g_i$ to pixel $p$. Instead of using a strict binary mask, we convert the score into a soft per-Gaussian coefficient that controls the strength of adversarial updates:
\begin{equation}
    m_i = \min\!\left(1,\left(\frac{s_i}{\tau_{\mathrm{mask}}+\epsilon}\right)^{\gamma_{\mathrm{mask}}}\right).
\end{equation}
Here, $\tau_{\mathrm{mask}}$ controls the score level at which the adversarial update scale saturates, while $\gamma_{\mathrm{mask}}$ controls how sharply this update coefficient decays for Gaussians with lower mask scores. In practice, $\mathcal{V}_{\mathrm{mask}}$ is sampled once before optimization, and the soft mask is kept fixed. The adversarial gradient for each Gaussian is scaled by $m_i$, so that mask-selected Gaussians receive stronger updates while surrounding Gaussians receive weaker updates. Rather than enforcing a hard separation between selected and surrounding Gaussians, the soft coefficient applies stronger adversarial updates to selected Gaussians while allowing weaker updates on nearby Gaussians. This soft allocation helps preserve rendering fidelity and watermark recovery by avoiding overly localized or overly aggressive perturbations. Experiments comparing hard and soft masking appear in Appendix~\ref{app:ablation}.

\subsubsection{Adversarial objective}
\label{subsec:adversarialobjective}
The adversarial branch aims to make the editing trajectory diverge from the intended one. For the sampled view set $\mathcal{V}_t$, we optimize an averaged multi-view objective:
\begin{equation}
     \mathcal{L}_{\mathrm{adv}} =
     \frac{1}{|\mathcal{V}_t|}
     \sum_{v \in \mathcal{V}_t}
     \Big( \mathcal{L}_{\mathrm{render}}^{(v)}
     - \lambda_{\mathrm{lat}}\,\mathcal{S}_{\mathrm{lat}}^{(v)}
     - \lambda_{\mathrm{traj}}\,\mathcal{S}_{\mathrm{traj}}^{(v)}
     - \lambda_{\mathrm{xattn}}\,\mathcal{S}_{\mathrm{xattn}}^{(v)} \Big).
    \label{eq:adv}
\end{equation}
Here, $\mathcal{L}_{\mathrm{render}}^{(v)}$ measures reconstruction error against the corresponding frozen reference render, $\lambda_{\mathrm{lat}}$, $\lambda_{\mathrm{traj}}$, and $\lambda_{\mathrm{xattn}}$ control the relative contributions of the three diversion terms, $\mathcal{G}^{\mathrm{ref}}$ denotes the frozen pretrained 3DGS scene, and $z(\cdot)$ is the VAE encoder. The latent-anchor separation term increases separation in the VAE latent space between the protected render and the frozen reference render:
\begin{equation}
    \mathcal{S}_{\mathrm{lat}}^{(v)}
    =
    \left\|
    z\big(\mathcal{R}_v(\mathcal{G}^{\mathrm{prot}})\big)
    -
    z\big(\mathcal{R}_v(\mathcal{G}^{\mathrm{ref}})\big)
    \right\|_2^2.
    \label{eq:latent-anchor}
\end{equation}

The denoising-trajectory diversion term, $\mathcal{S}_{\mathrm{traj}}^{(v)}$, separates the protected render from the frozen reference render under the same edit prompt $\psi$, diffusion timestep $t$, and sampled noise $\epsilon$~\citep{nulltextinversion}. Let $\Phi_{\mathrm{traj}}(I;\psi,t,\epsilon)$ denote the frozen editor's denoising-trajectory descriptor, which summarizes the prompt-induced edit direction and a short denoising trajectory from image $I$. We define
\begin{equation}
    \mathcal{S}_{\mathrm{traj}}^{(v)}
    =
    \left\|
    \Phi_{\mathrm{traj}}\!\left(\mathcal{R}_v(\mathcal{G}^{\mathrm{prot}});\psi,t,\epsilon\right)
    -
    \Phi_{\mathrm{traj}}\!\left(\mathcal{R}_v(\mathcal{G}^{\mathrm{ref}});\psi,t,\epsilon\right)
    \right\|_2^2.
    \label{eq:trajectory-diversion}
\end{equation}
This term pushes the protected render away from the reference editing direction even when the protected and reference renders remain visually similar before editing.

We additionally use a cross-attention diversion term, $\mathcal{S}_{\mathrm{xattn}}^{(v)}$, which enlarges the query separation in the mid-block text-conditioned cross-attention of the frozen diffusion editor~\citep{prompttoprompt}:
\begin{equation}
    \mathcal{S}_{\mathrm{xattn}}^{(v)}
    =
    \frac{1}{d_{\mathrm{mid}}}
    \left\|
    \Phi_{\mathrm{mid}}^{\mathrm{xattn}}\!\left(\mathcal{R}_v(\mathcal{G}^{\mathrm{prot}});\psi,t,\epsilon\right)
    -
    \Phi_{\mathrm{mid}}^{\mathrm{xattn}}\!\left(\mathcal{R}_v(\mathcal{G}^{\mathrm{ref}});\psi,t,\epsilon\right)
    \right\|_2^2,
    \label{eq:xattn-main}
\end{equation}
where $\Phi_{\mathrm{mid}}^{\mathrm{xattn}}$ denotes a descriptor obtained by pooling query tokens from the mid-block text-conditioned cross-attention. By increasing this diversion term, the protected render induces prompt-to-image alignment that differs from that of the reference, disrupting the intended editing trajectory. For a detailed definition of the cross-attention diversion term in Eq.~\eqref{eq:xattn-main}, please refer to Appendix~\ref{app:xattn}.

\subsubsection{Parameter-role-aware adversarial modulation}
\label{subsec:roleawaremodulation}
The update-saliency-motivated Gaussian selection determines which Gaussians should receive stronger adversarial signals. However, it does not resolve another source of imbalance: each Gaussian consists of parameters with different rendering roles whose perturbations have different effects on rendered images. In particular, the scale $s_i$ and rotation $r_i$ determine the anisotropic covariance of a Gaussian, opacity $\alpha_i$ controls visibility and alpha compositing, the DC color coefficient $\mathbf{c}^{\mathrm{dc}}_i$ controls view-independent appearance, and the remaining spherical-harmonic coefficients $\mathbf{c}^{\mathrm{rest}}_i$ control view-dependent residual appearance. Therefore, applying the same adversarial gradient scale to all parameter groups does not reliably control the resulting rendered-image degradation.

To account for the different rendering sensitivities of Gaussian parameter groups, we assign a parameter-role coefficient $\rho_k$ to each parameter group $k$ and modulate only the adversarial gradient:
\begin{equation}
    \widetilde{\nabla}_{\theta_{i,k}}\mathcal{L}_{\mathrm{adv}}
    =
    m_i\,\rho_k\,
    \nabla_{\theta_{i,k}}\mathcal{L}_{\mathrm{adv}},
    \label{eq:role-aware-grad}
\end{equation}
where $\theta_{i,k}$ denotes the $k$-th parameter group of Gaussian $g_i$. The final gradient $\nabla_{\theta_{i,k}}\mathcal{L}$ used to update the protected scene combines the watermark gradient with the modulated adversarial gradient:
\begin{equation}
    \nabla_{\theta_{i,k}}\mathcal{L}
    =
    \nabla_{\theta_{i,k}}\mathcal{L}_{\mathrm{wm}}
    +
    \lambda_{\mathrm{adv}}\,
    \widetilde{\nabla}_{\theta_{i,k}}\mathcal{L}_{\mathrm{adv}}.
    \label{eq:role-aware-total-grad}
\end{equation}
The parameter-role coefficients are chosen to reflect the rendering sensitivity of each parameter role: opacity and DC color are treated more conservatively because they can cause visible density or color drift across views, whereas scale, rotation, and view-dependent residual features can carry adversarial signals with less direct global appearance shift.
\begin{table*}[t]
    \centering
    \caption{\textbf{Quantitative results.} The results are averaged over the benchmark set. PSNR, SSIM, LPIPS, and bit accuracy are measured from rendered images. CLIP, CLIP-T, and CLIP-D are measured after 3D editing using DGE. 3DGSW+DEGauss denotes a straightforward composition of watermarking and adversarial objectives. Orig. denotes the value after editing a pretrained 3DGS. GuardMarkGS achieves the highest sUCPS, which jointly evaluates traceability, edit deterrence, and fidelity.}
    \label{tab:main}
    \small
    \vspace{-5pt}
    \renewcommand{\arraystretch}{1.25} 
    \resizebox{\textwidth}{!}{%
    \begin{tabular}{lcccccccccccccc}
        \toprule
        \multirow{2}{*}{Method} & \multirow{2}{*}{sUCPS $\uparrow$} & \multirow{2}{*}{Bit Acc. (\%) $\uparrow$} & \multicolumn{3}{c}{CLIP} & \multicolumn{3}{c}{CLIP-T} & \multicolumn{3}{c}{CLIP-D} & \multirow{2}{*}{PSNR $\uparrow$} & \multirow{2}{*}{SSIM $\uparrow$} & \multirow{2}{*}{LPIPS $\downarrow$} \\
        \cmidrule(lr){4-6} \cmidrule(lr){7-9} \cmidrule(lr){10-12}
         &  &  & Orig. & Method $\downarrow$ & Diff. $\uparrow$ & Orig. & Method $\downarrow$ & Diff. $\uparrow$ & Orig. & Method $\downarrow$ & Diff. $\uparrow$ &  &  &  \\
        \midrule
        3DGSW~\citep{3dgsw} 
        & \underline{0.7791} & \textbf{99.00} 
        & \multirow{6}{*}{1.0000} & 0.9629 & 0.0371 
        & \multirow{6}{*}{0.2825} & 0.2826 & -0.0001 
        & \multirow{6}{*}{0.1476} & 0.1408 & 0.0068 
        & \underline{33.94} & \underline{0.9505} & 0.0866 \\
        
        GaussianMarker~\citep{gaussianmarker} 
        & 0.7516 & 98.51 
        &  & 0.9731 & 0.0269 
        &  & 0.2835 & -0.0010 
        &  & 0.1537 & -0.0061 
        & \textbf{35.37} & \textbf{0.9710} & \textbf{0.0597} \\
        
        GuardSplat~\citep{guardsplt} 
        & 0.7489 & \underline{98.92} 
        &  & 0.9854 & 0.0146 
        &  & 0.2819 & 0.0006 
        &  & 0.1492 & -0.0016 
        & 29.34 & 0.9225 & \underline{0.0627} \\
        
        DEGauss~\citep{degauss} 
        & 0.6467 & N/A 
        &  & \underline{0.9250} & \underline{0.0750} 
        &  & \underline{0.2713} & \underline{0.0112} 
        &  & \underline{0.1268} & \underline{0.0208} 
        & 30.34 & 0.9120 & 0.1502 \\
        
        3DGSW+DEGauss 
        & 0.6200 & 62.79 
        &  & 0.9596 & 0.0404 
        &  & 0.2813 & 0.0012 
        &  & 0.1387 & 0.0089 
        & 29.90 & 0.8463 & 0.1950 \\
        
        \rowcolor{gray!15}
        GuardMarkGS (Ours) 
        & \textbf{0.8622} & 97.23 
        &  & \textbf{0.9093} & \textbf{0.0907} 
        &  & \textbf{0.2676} & \textbf{0.0149} 
        &  & \textbf{0.1086} & \textbf{0.0390} 
        & 30.36 & 0.8935 & 0.1471 \\
        \bottomrule
    \end{tabular}
    }
\vspace{-10pt}
\end{table*}
\begin{figure*}[t]
    \centering
    \begin{minipage}{\textwidth}
        \centering
        \includegraphics[width=\linewidth]{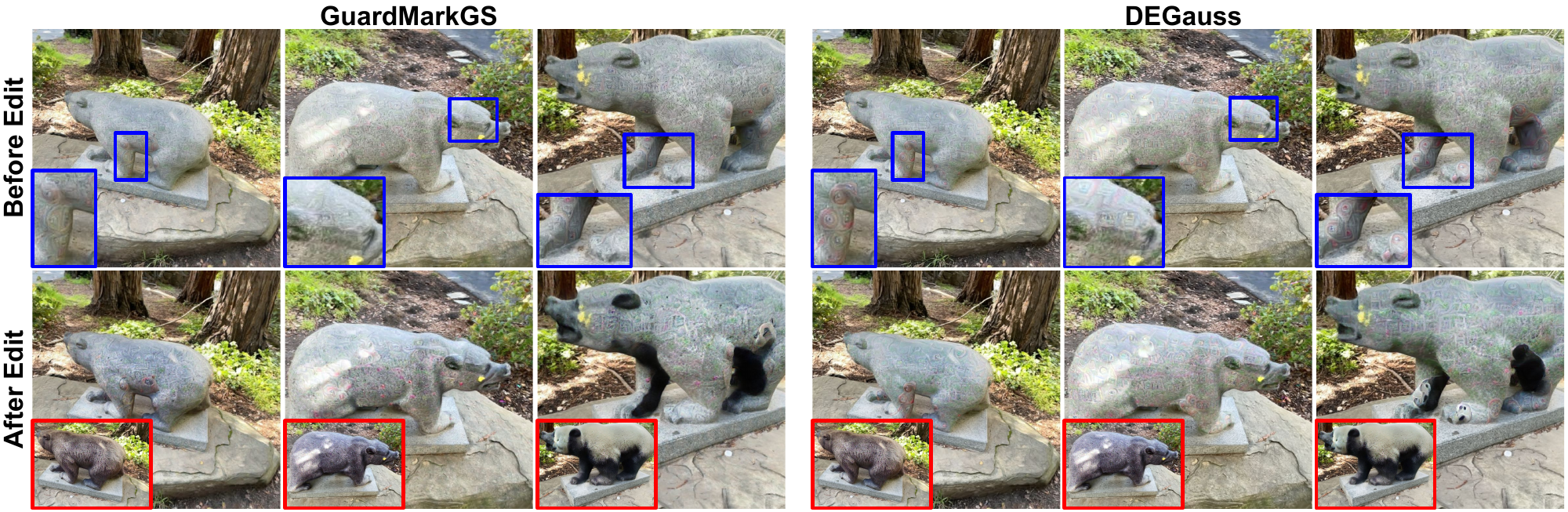}
        \caption{\textbf{Qualitative results.} We compare rendered images before and after 3D editing using DGE. To qualitatively evaluate editing defense performance, we compare our method with DEGauss~\citep{degauss}. The first row shows pre-edit renders, while the second row presents the corresponding post-edit outputs. Here, the \textit{red boxes} indicate the edited results of pretrained 3DGS. Despite providing both traceability and edit deterrence, our method maintains higher rendering fidelity than DEGauss.}
        \label{fig:qualitative}
    \end{minipage}
\vspace{-15pt}
\end{figure*}

\section{Experiments}
\label{sec:experiments}
\subsection{Setups}
\label{subsec:experimentalsetup}
\textbf{Scenes and datasets.} We evaluate on \textit{bicycle} and \textit{garden} from Mip-NeRF 360~\citep{mipnerf360}, and \textit{face}, \textit{person}, \textit{bear}, and \textit{fangzhou} from Instruct-NeRF2NeRF~\citep{nerf2nerf}. All evaluations are conducted on novel views, so that our method and the baselines are compared fairly under the same rendering setting.

\textbf{Baselines.} We compare against watermarking baselines, 3DGSW~\citep{3dgsw}, GaussianMarker~\citep{gaussianmarker}, and GuardSplat~\citep{guardsplt}, as well as one defense-only baseline, DEGauss~\citep{degauss}. To examine whether joint protection can be obtained by a simple composition of existing methods, we include 3DGSW+DEGauss, which adds DEGauss-style adversarial terms to a 3DGSW-style watermarking objective.

\textbf{Editing pipelines.} We use DGE~\citep{dge} as the main unauthorized editing pipeline, as it is a representative 3DGS editing method enforcing multi-view consistency. We evaluate defense performance using GaussianEditor~\citep{gaussianeditor} to examine whether edit deterrence transfers to another editing pipeline.

\textbf{Evaluation metrics.} We evaluate our method and the baselines from three perspectives. \textit{Traceability}: we render images from novel views and measure watermark bit accuracy from the decoded ownership message. \textit{Deterrence capability}: after applying the editing pipelines, we render the edited scenes from novel views and measure CLIP~\citep{clip}, CLIP-T, and CLIP-D. Here, CLIP measures the similarity between the edited results from the protected and original scenes, CLIP-T measures the similarity between the edited protected result and the target text prompt, and CLIP-D measures how strongly the image editing direction aligns with the text editing direction. \textit{Fidelity}: after applying each protection method to a pretrained 3DGS model, we compare the rendered protected images against the original renders using PSNR, SSIM~\citep{ssim}, and LPIPS~\citep{lpips}. To assess the balance across these three aspects, we define a unified score, \textit{soft Unified Copyright Protection Score} (sUCPS), which combines traceability, edit deterrence, and fidelity after per-metric normalization. This metric is high only when consistently superior performance is demonstrated across all aspects. Please refer to Appendix~\ref{app:clip_metrics} for the definitions of CLIP, CLIP-T, and CLIP-D, and to Appendix~\ref{app:sucps} for the definition of sUCPS.

\textbf{Implementation details.} Experiments use an NVIDIA RTX A6000 GPU. Each scene is optimized for 8 epochs using a 32-bit watermark message, and target Gaussians for adversarial noise injection are selected using 2D masks from 24 viewpoints. We set $\lambda_{\mathrm{msg}}=0.1$, $\lambda_{\mathrm{quality}}=1.0$, and $\lambda_{\mathrm{lat}}=\lambda_{\mathrm{traj}}=\lambda_{\mathrm{xattn}}=10^{-4}$. Additional algorithmic details and edit prompts are provided in Appendix~\ref{app:algorithm}.
\begin{table*}[t]
    \centering
    \caption{\textbf{Ablation study of the joint optimization design.} Since watermarking and adversarial objectives are essential for joint optimization, we omit them from the columns and present ablation studies on update-saliency-motivated Gaussian selection and parameter-role-aware adversarial modulation. Results show that the proposed components improve the trade-off among traceability, edit deterrence, and rendering fidelity, with the full model achieving the best balance.}
    \label{tab:joint-design-ablation}
    \small
    \vspace{-5pt}
    \renewcommand{\arraystretch}{1.25}
    \resizebox{\textwidth}{!}{%
    \begin{tabular}{cccccccccccccccc}
        \toprule
        \multirow{2}{*}{\shortstack[c]{Gaussian\\Selection}} 
        & \multirow{2}{*}{\shortstack[c]{Adversarial\\Modulation}} 
        & \multirow{2}{*}{sUCPS $\uparrow$}
        & \multirow{2}{*}{Bit Acc. (\%) $\uparrow$} 
        & \multicolumn{3}{c}{CLIP} 
        & \multicolumn{3}{c}{CLIP-T} 
        & \multicolumn{3}{c}{CLIP-D} 
        & \multirow{2}{*}{PSNR $\uparrow$} 
        & \multirow{2}{*}{SSIM $\uparrow$} 
        & \multirow{2}{*}{LPIPS $\downarrow$} \\
        \cmidrule(lr){5-7} \cmidrule(lr){8-10} \cmidrule(lr){11-13}
        & & & 
        & Orig. & Method $\downarrow$ & Diff. $\uparrow$
        & Orig. & Method $\downarrow$ & Diff. $\uparrow$
        & Orig. & Method $\downarrow$ & Diff. $\uparrow$
        & & & \\
        \midrule
        - & - 
        & 0.6776
        & 67.80
        & \multirow{4}{*}{1.0000} & \textbf{0.8706} & \textbf{0.1294}
        & \multirow{4}{*}{0.2825} & \textbf{0.2620} & \textbf{0.0205}
        & \multirow{4}{*}{0.1476} & \textbf{0.0851} & \textbf{0.0625}
        & 28.73 & 0.8159 & 0.3069 \\

        \checkmark & - 
        & \underline{0.8566}
        & \underline{97.02} 
        &  & 0.9088 & 0.0912
        &  & 0.2678 & 0.0147
        &  & 0.1098 & 0.0378
        & \underline{30.22} & \underline{0.8801} & \underline{0.1503} \\

        - & \checkmark 
        & 0.6683
        & 65.02 
        &  & \underline{0.8734} & \underline{0.1266}
        &  & \underline{0.2625} & \underline{0.0200}
        &  & \underline{0.0890} & \underline{0.0586}
        & 28.74 & 0.8161 & 0.3062 \\

        \rowcolor{gray!15}
        \checkmark & \checkmark 
        & \textbf{0.8622}
        & \textbf{97.23} 
        &  & 0.9093 & 0.0907
        &  & 0.2676 & 0.0149
        &  & 0.1086 & 0.0390
        & \textbf{30.36} & \textbf{0.8935} & \textbf{0.1471} \\
        \bottomrule
    \end{tabular}
    }
\vspace{-10pt}
\end{table*}

\subsection{Main results}
\label{subsec:experimentalresults}
\textbf{Quantitative results.} Tab.~\ref{tab:main} summarizes the averaged quantitative results over the evaluated benchmark set, including the proposed unified score sUCPS. The watermark-only methods achieve high bit accuracy and strong rendering fidelity, but they show clear limitations on the edit-deterrence metrics. DEGauss is included as a defense-only reference, but it does not provide watermark decoding and therefore cannot satisfy the ownership-tracing requirement. The 3DGSW+DEGauss further tests a straightforward composition of existing watermarking and edit-deterrence objectives. This naive combination fails to achieve a stable balance, as it substantially reduces bit accuracy and rendering quality while providing only limited edit-deterrence gains. These results show that joint protection cannot be obtained by simply adding adversarial edit-deterrence terms to a watermarking objective. In contrast, our method demonstrates recoverable ownership tracing with high bit accuracy and effective editing defense, while still maintaining competitive fidelity. Taken together, these results suggest that GuardMarkGS provides a more integrated and practical solution for copyright protection of 3DGS-based assets by jointly addressing owner identification and unauthorized 3D edit deterrence.

\textbf{Qualitative results.} Fig.~\ref{fig:qualitative} compares our method with DEGauss~\citep{degauss} before and after 3D editing using DGE~\citep{dge}. The defense-only baseline introduces visible degradation in the pre-edit renders, whereas our method maintains higher rendering fidelity while still disrupting the edited results. These qualitative results indicate that our method provides a better balance between fidelity and edit deterrence, in addition to supporting ownership traceability through watermark recovery. Additional qualitative results are provided in Appendix~\ref{app:qualitative}.

\textbf{Additional generalization experiments.} We further evaluate edit deterrence under GaussianEditor~\citep{gaussianeditor}, a 3D editing pipeline other than DGE~\citep{dge}. The results show that the proposed method is not limited to a single editing method and can still suppress unauthorized edits under this alternative editor. The results for this cross-editor generalization setting are provided in Appendix~\ref{app:generalizability}.

\subsection{Ablation study}
\label{subsec:ablation}
\textbf{Effect of update-saliency-motivated selection.} Tab.~\ref{tab:joint-design-ablation} reports the ablation of update-saliency-motivated Gaussian selection under the same scene-wide watermarking and adversarial objectives. When parameter-role modulation is enabled, adding the proposed selection strategy improves bit accuracy from $65.02$ to $97.23$, PSNR from $28.74$ to $30.36$, and SSIM from $0.8161$ to $0.8935$, while reducing LPIPS from $0.3062$ to $0.1471$. Although the variants without selection can produce stronger raw CLIP-based edit-deterrence gaps, they do so with severe degradation in watermark recovery and rendering fidelity. This indicates that Gaussian selection mainly stabilizes the joint optimization. Additional results comparing hard and soft Gaussian masking are provided in Appendix~\ref{app:ablation}.
\begin{table*}[t]
    \centering
    \caption{\textbf{Ablation on adversarial diversion-term combinations.} We evaluate various combinations of latent-anchor separation $\mathcal{S}_{\mathrm{lat}}$, denoising-trajectory diversion $\mathcal{S}_{\mathrm{traj}}$, and cross-attention diversion $\mathcal{S}_{\mathrm{xattn}}$. The results show that the three terms work complementarily to achieve strong edit deterrence.}
    \label{tab:loss-combination}
    \small
    \vspace{-5pt}
    \renewcommand{\arraystretch}{1.25}
    \resizebox{\textwidth}{!}{%
    \begin{tabular}{ccccccccccccccccc}
        \toprule
        \multirow{2}{*}{$\mathcal{S}_{\mathrm{lat}}$} 
        & \multirow{2}{*}{$\mathcal{S}_{\mathrm{traj}}$} 
        & \multirow{2}{*}{$\mathcal{S}_{\mathrm{xattn}}$} 
        & \multirow{2}{*}{sUCPS $\uparrow$}
        & \multirow{2}{*}{Bit Acc. (\%) $\uparrow$} 
        & \multicolumn{3}{c}{CLIP} 
        & \multicolumn{3}{c}{CLIP-T} 
        & \multicolumn{3}{c}{CLIP-D} 
        & \multirow{2}{*}{PSNR $\uparrow$} 
        & \multirow{2}{*}{SSIM $\uparrow$} 
        & \multirow{2}{*}{LPIPS $\downarrow$} \\
        \cmidrule(lr){6-8} \cmidrule(lr){9-11} \cmidrule(lr){12-14}
        & & & & 
        & Orig. & Method $\downarrow$ & Diff. $\uparrow$
        & Orig. & Method $\downarrow$ & Diff. $\uparrow$
        & Orig. & Method $\downarrow$ & Diff. $\uparrow$
        & & & \\
        \midrule
        - & \checkmark & \checkmark 
        & \underline{0.8621}
        & 95.35 
        & \multirow{7}{*}{1.0000} & 0.9251 & 0.0749
        & \multirow{7}{*}{0.2825} & \underline{0.2712} & \underline{0.0113}
        & \multirow{7}{*}{0.1476} & \underline{0.1232} & \underline{0.0244}
        & 30.02 & 0.8956 & 0.1398 \\

        \checkmark & - & \checkmark 
        & 0.7733
        & \underline{97.12} 
        &  & 0.9562 & 0.0438
        &  & 0.2795 & 0.0030
        &  & 0.1426 & 0.0050
        & \underline{31.24} & 0.9022 & 0.1301 \\

        \checkmark & \checkmark & - 
        & 0.8595
        & 95.42
        &  & \underline{0.9224} & \underline{0.0776}
        &  & 0.2737 & 0.0088
        &  & 0.1237 & 0.0239
        & 30.32 & 0.8944 & 0.1333 \\

        - & - & \checkmark 
        & 0.7686
        & 96.59 
        &  & 0.9642 & 0.0358
        &  & 0.2798 & 0.0027
        &  & 0.1412 & 0.0064
        & \textbf{31.28} & \underline{0.9043} & 0.1298 \\

        \checkmark & - & - 
        & 0.7325
        & \underline{97.12} 
        &  & 0.9644 & 0.0356
        &  & 0.2814 & 0.0011
        &  & 0.1474 & 0.0002
        & 31.20 & \textbf{0.9067} & \underline{0.1288} \\

        - & \checkmark & - 
        & 0.8402
        & 95.13 
        &  & 0.9306 & 0.0694
        &  & 0.2747 & 0.0078
        &  & 0.1334 & 0.0142
        & 30.22 & 0.8999 & \textbf{0.1286} \\

        \rowcolor{gray!15}
        \checkmark & \checkmark & \checkmark 
        & \textbf{0.8622}
        & \textbf{97.23} 
        &  & \textbf{0.9093} & \textbf{0.0907}
        &  & \textbf{0.2676} & \textbf{0.0149}
        &  & \textbf{0.1086} & \textbf{0.0390}
        & 30.36 & 0.8935 & 0.1471 \\
        \bottomrule
    \end{tabular}
    }
\vspace{-10pt}
\end{table*}

\textbf{Effect of parameter-role modulation.} Tab.~\ref{tab:joint-design-ablation} also assesses the effect of the role-aware coefficients in Eq.~\eqref{eq:role-aware-grad}. Fixing the proposed Gaussian selection, adding parameter-role-aware modulation improves sUCPS from $0.8566$ to $0.8622$, bit accuracy from $97.02$ to $97.23$, and PSNR from $30.22$ to $30.36$. The CLIP-based edit-deterrence scores remain comparable, with CLIP changing from $0.9088$ to $0.9093$, CLIP-T from $0.2678$ to $0.2676$, and CLIP-D from $0.1098$ to $0.1086$. These results suggest that parameter-role-aware modulation further improves the fidelity side of the trade-off while preserving the edit-deterrence effect of the selected adversarial updates. Additional ablations on the perturbed Gaussian parameter groups are provided in Appendix~\ref{app:param_group}.

\textbf{Diversion-term analysis.} We analyze how the three diversion terms in Eq.~\eqref{eq:adv} contribute to editing defense performance. Specifically, we examine latent-anchor separation $\mathcal{S}_{\mathrm{lat}}$ in Eq.~\eqref{eq:latent-anchor}, denoising-trajectory diversion $\mathcal{S}_{\mathrm{traj}}$ in Eq.~\eqref{eq:trajectory-diversion}, and cross-attention diversion $\mathcal{S}_{\mathrm{xattn}}$ in Eq.~\eqref{eq:xattn-main}. Tab.~\ref{tab:loss-combination} shows that rendering fidelity and bit accuracy remain comparable across variants, whereas edit-deterrence metrics vary more substantially. This indicates that the performance gain does not simply come from injecting a larger amount of adversarial noise, but from the complementary effects of the three diversion terms. Using all three terms achieves stronger editing defense, resulting in a better balance across traceability, edit deterrence, and rendering fidelity, yielding the highest sUCPS.

\textbf{Ablation of the joint protection design.} Tab.~\ref{tab:joint-design-ablation} summarizes how the two allocation mechanisms interact under the fixed watermarking and adversarial objectives. Without both Gaussian selection and role-aware modulation, the model produces strong raw CLIP-based edit-deterrence gaps, but suffers from low bit accuracy and poor rendering fidelity, indicating a severe conflict between the two protection signals. Gaussian selection provides the main improvement in watermark recovery and rendering fidelity by controlling the spatial support of adversarial updates. Role-aware modulation further refines this allocation by scaling adversarial gradients according to Gaussian parameter roles. The full model achieves the best overall sUCPS, suggesting that effective joint protection requires explicit control over both the selected Gaussian support and the parameter-role sensitivity, rather than simply applying watermarking and adversarial objectives to the same 3DGS support.

\section{Conclusion}
\label{sec:conclusion}
We present a unified protection framework for 3D Gaussian Splatting that jointly addresses ownership tracing and unauthorized edit deterrence. The core idea is to embed the watermark signal over the full scene while assigning stronger adversarial updates to a mask-selected Gaussian subset. This design addresses a central tension in 3DGS copyright protection: ownership evidence should remain recoverable across diverse viewpoints through scene-wide watermarking, while adversarial signals should be selectively allocated under joint optimization rather than uniformly imposed on the same scene-wide support. This allocation helps maintain a stable balance among traceability, edit deterrence, and rendering fidelity. The experimental results show that the proposed framework achieves the best sUCPS, our unified score that jointly reflects traceability, edit deterrence, and fidelity, while maintaining competitive performance across all three aspects. We believe that future 3D copyright protection systems should move beyond isolated objectives, such as passive ownership tracing or active edit deterrence, and instead optimize them jointly.

\textbf{Limitations and future works.} We have shown that copyright tracing via digital watermarking and active copyright protection through adversarial signal injection can be achieved simultaneously within a unified framework. However, to become a practical copyright protection technique, the method must demonstrate sufficient robustness under a wider range of distortion scenarios, including not only distortions applied to rendered images but also model-level perturbations that directly modify the protected 3D representation itself. Therefore, future work should include a broader and more systematic analysis under realistic copyright-protection scenarios that incorporate such distortions.



\clearpage
{\small
\bibliographystyle{unsrtnat}
\bibliography{neurips_2026}
}


\newpage
\appendix
\section*{Appendix}
\section{Broader impact}
The proposed framework can help creators and organizations protect 3D assets in digital heritage, virtual production, AR/VR, and simulation pipelines. By combining ownership tracing and edit deterrence within a protected 3DGS representation, our method supports practical copyright protection even when released assets are redistributed or subjected to instruction-driven editing. This can help content owners preserve ownership evidence, discourage unauthorized manipulation, and deploy protected 3D assets while maintaining rendering fidelity. More broadly, our work highlights the need for copyright protection systems that jointly consider traceability, edit deterrence, and visual quality, providing a foundation for future research on reliable protection of editable 3D representations.

\section{Additional algorithmic details}
\label{app:algorithm}
In this section, we provide the algorithmic details mentioned in Sec.~\ref{sec:experiments} of the main paper. The overall optimization procedure is summarized in Alg.~\ref{alg:guardmarkgs_training}. In addition, Tab.~\ref{tab:edit-prompts} lists the edit text prompts used to evaluate edit-deterrence performance, including the main evaluation reported in Tab.~\ref{tab:main}.
\begin{algorithm}[ht]
\caption{Joint optimization of GuardMarkGS}
\label{alg:guardmarkgs_training}
\begin{algorithmic}[1]
\Require Pretrained 3DGS scene $\mathcal{G}$, frozen reference scene $\mathcal{G}^{\mathrm{ref}}\leftarrow\mathcal{G}$, training view set $\mathcal{V}$, ownership message $\mathbf{m}$, prompt library $\Psi_{\mathrm{lib}}$, parameter-role coefficients $\{\rho_k\}$, total iterations $T$
\Ensure Protected 3DGS scene $\mathcal{G}^{\mathrm{prot}}$

\State Initialize $\mathcal{G}^{\mathrm{prot}} \leftarrow \mathcal{G}$
\State Precompute a soft Gaussian mask $\{m_i\}$ from text-guided multi-view semantic segmentation
\State Set role coefficients $\{\rho_k\}$ for Gaussian parameter groups

\For{$t = 1$ to $T$}
    \State Sample a multi-view set $\mathcal{V}_t \subset \mathcal{V}$
    \State Sample an edit prompt $\psi \sim \Psi_{\mathrm{lib}}$

    \For{each $v \in \mathcal{V}_t$}
        \State Render protected view $I_v^{\mathrm{prot}}=\mathcal{R}_v(\mathcal{G}^{\mathrm{prot}})$
        \State Render reference view $I_v^{\mathrm{ref}}=\mathcal{R}_v(\mathcal{G}^{\mathrm{ref}})$
        \State Compute watermark loss $\mathcal{L}_{\mathrm{wm}}^{(v)}$
        \State Compute adversarial loss $\mathcal{L}_{\mathrm{adv}}^{(v)}$
    \EndFor

    \State Aggregate the scene-wide watermark loss over $\mathcal{V}_t$
    \[
        \mathcal{L}_{\mathrm{wm}}
        =
        \frac{1}{|\mathcal{V}_t|}
        \sum_{v\in\mathcal{V}_t}
        \mathcal{L}_{\mathrm{wm}}^{(v)}
    \]
    \State Aggregate the averaged adversarial loss over $\mathcal{V}_t$
    \[
        \mathcal{L}_{\mathrm{adv}}
        =
        \frac{1}{|\mathcal{V}_t|}
        \sum_{v\in\mathcal{V}_t}
        \mathcal{L}_{\mathrm{adv}}^{(v)}
    \]
    \State Backpropagate $\mathcal{L}_{\mathrm{wm}}$ to obtain scene-wide watermark gradients
    \State Compute adversarial gradients $\nabla_{\theta_{i,k}}\mathcal{L}_{\mathrm{adv}}$
    \State Apply the soft Gaussian mask and role coefficients:
    \[
        \widetilde{\nabla}_{\theta_{i,k}}\mathcal{L}_{\mathrm{adv}}
        =
        m_i\rho_k\nabla_{\theta_{i,k}}\mathcal{L}_{\mathrm{adv}}
    \]
    \State Combine the watermark and $\lambda_{\mathrm{adv}}$-weighted modulated adversarial gradients
    \State Update the parameters of $\mathcal{G}^{\mathrm{prot}}$ with the optimizer
\EndFor

\State \Return $\mathcal{G}^{\mathrm{prot}}$
\end{algorithmic}
\end{algorithm}
\begin{table}[ht]
    \centering
    \caption{\textbf{Edit prompts used for evaluating edit-deterrence performance.} For each evaluated scene, we list the text prompts used to perform 3D editing under each scene-specific evaluation setting.}
    \label{tab:edit-prompts}
    \small
    \setlength{\tabcolsep}{6pt}
    \renewcommand{\arraystretch}{1.2}
    \newcolumntype{Y}{>{\raggedright\arraybackslash}X}
    \begin{tabularx}{\linewidth}{c Y c Y}
        \toprule
        \textbf{scene} & \textbf{edit text prompt} & \textbf{scene} & \textbf{edit text prompt} \\
        \midrule

        bear &
        Turn the bear statue into a panda \newline
        Turn the bear statue into a grizzly bear \newline
        Turn the bear statue into an asiatic black bear \newline
        Turn the bear statue into a wild boar
        &
        bicycle &
        Change the bicycle color to bright red \newline
        Turn the ground into a Namibian desert \newline
        A photo of the bicycle at the Namibian desert
        \\
        \midrule

        face &
        Wear him a glasses \newline
        Make him wear a Venetian mask \newline
        Turn him into the Tolkien Elf \newline
        Turn him into an Einstein \newline
        A photo of a marble sculpture
        &
        garden &
        Change the table color to deep mahogany brown \newline
        Turn the vase into red \newline
        A photo of a garden scene with autumn
        \\
        \midrule

        person &
        Make the man look like a mosaic Sculpture \newline
        Turn the man into a robot \newline
        Turn him into a Minecraft character \newline
        Make him reading a book
        &
        fangzhou &
        Turn him into a comic book character \newline
        Turn him into an old man with wrinkles \newline
        Turn him into a spider man with Mask
        \\

        \bottomrule
    \end{tabularx}
\vspace{-5pt}
\end{table}

\section{Additional results under different editing pipeline}
\label{app:generalizability}
\begin{figure*}[ht]
    \centering
    \begin{minipage}{\textwidth}
        \centering
        \includegraphics[width=\linewidth]{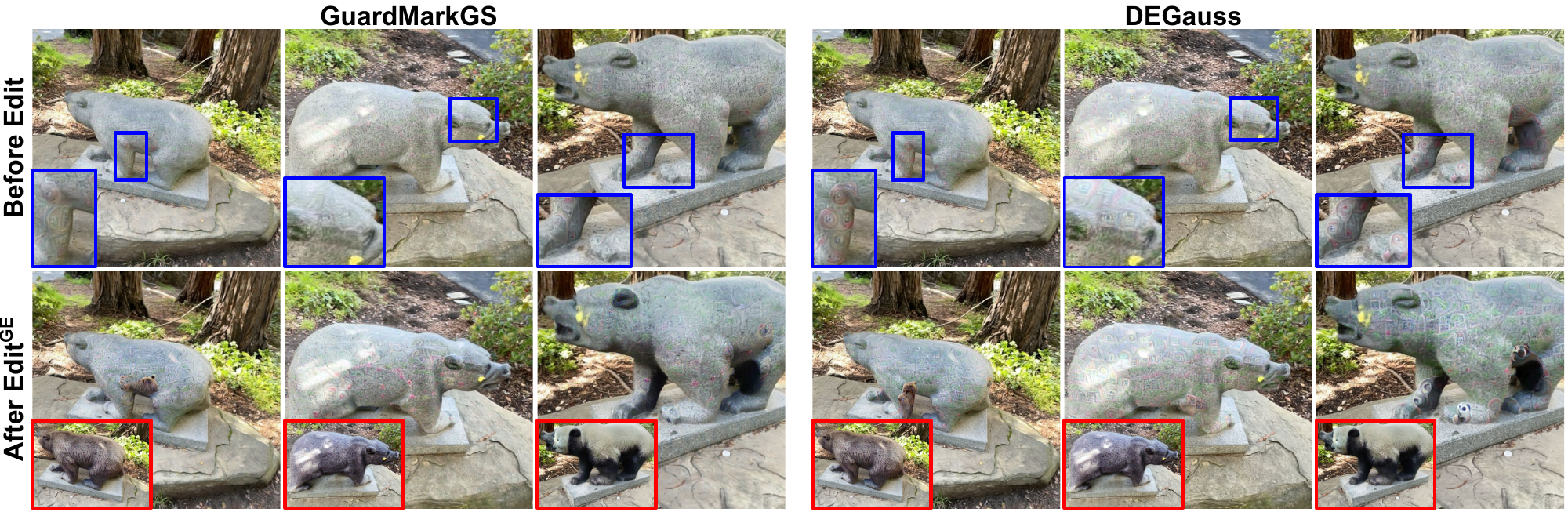}
        \caption{\textbf{Generalization to GaussianEditor: qualitative results.} We show qualitative results on GaussianEditor~\citep{gaussianeditor}, a 3D editing method other than DGE~\citep{dge}. \textit{After Edit$^{\mathrm{GE}}$} denotes results edited with GaussianEditor. The \textit{red boxes} in the second row show edited results of pretrained 3DGS. These results indicate our method's edit-deterrence is not limited to the main DGE pipeline.}
        \label{fig:generalizability}
    \end{minipage}
\vspace{-15pt}
\end{figure*}

We provide additional quantitative results using GaussianEditor~\citep{gaussianeditor}, a 3D editing pipeline other than DGE~\citep{dge}. Tab.~\ref{tab:generalability_gaussianeditor} reports edit-deterrence results when GaussianEditor is used. In addition, Fig.~\ref{fig:generalizability} shows qualitative results edited with GaussianEditor. Taken together, these results show that the edit-deterrence performance of our proposed method is not limited to a specific editing model.
\begin{table*}[ht]
    \centering
    \caption{\textbf{Generalization to GaussianEditor: quantitative results.} Lower CLIP, CLIP-T, and CLIP-D values indicate more successful editing defense. Here, the Orig. column reports the values measured from edits of the original scene, while the Method column reports the values measured from edits of the protected scene produced by each method. GuardMarkGS also achieves the strongest editing defense performance in experiments using GaussianEditor~\citep{gaussianeditor}.}
    \label{tab:generalability_gaussianeditor}
    \small
    \vspace{-5pt}
    \renewcommand{\arraystretch}{1.10}
    \resizebox{\textwidth}{!}{%
    \begin{tabular}{lccccccccc}
        \toprule
        \multirow{2}{*}{Method} & \multicolumn{3}{c}{CLIP} & \multicolumn{3}{c}{CLIP-T} & \multicolumn{3}{c}{CLIP-D} \\
        \cmidrule(lr){2-4} \cmidrule(lr){5-7} \cmidrule(lr){8-10}
         & Orig. & Method $\downarrow$ & Diff. $\uparrow$ & Orig. & Method $\downarrow$ & Diff. $\uparrow$ & Orig. & Method $\downarrow$ & Diff. $\uparrow$ \\
        \midrule
        3DGSW~\citep{3dgsw} 
        & \multirow{5}{*}{1.0000} & 0.9582 & 0.0418 
        & \multirow{5}{*}{0.2737} & 0.2770 & -0.0033 
        & \multirow{5}{*}{0.1293} & 0.1306 & -0.0013 \\
        
        GaussianMarker~\citep{gaussianmarker} 
        &  & 0.9643 & 0.0357 
        &  & 0.2779 & -0.0042 
        &  & 0.1449 & -0.0156 \\
        
        GuardSplat~\citep{guardsplt} 
        &  & 0.9777 & 0.0223 
        &  & 0.2754 & -0.0017 
        &  & 0.1362 & -0.0069 \\
        
        DEGauss~\citep{degauss} 
        &  & \underline{0.9274} & \underline{0.0726} 
        &  & \underline{0.2711} & \underline{0.0026} 
        &  & \underline{0.1282} & \underline{0.0011} \\
        
        \rowcolor{gray!15}
        GuardMarkGS (Ours) 
        &  & \textbf{0.9243} & \textbf{0.0757}
        &  & \textbf{0.2696} & \textbf{0.0041}
        &  & \textbf{0.1196} & \textbf{0.0097} \\
        \bottomrule
    \end{tabular}
    }
\end{table*}
\vspace{-15pt}

\section{Additional robustness experiments}
\label{app:robustness}
In the main paper, Tab.~\ref{tab:main} reports the primary quantitative comparison results. In this section, we present additional experiments to further examine the robustness of the proposed framework. Here, robustness refers to the ability to preserve both ownership tracing and defense against 3D editing even under scenarios where copyright-protection mechanisms may be weakened.

\textbf{Robustness of the watermark.} In the application scenario of Fig.~\ref{fig:teaser}, Alice verifies ownership by recovering the embedded watermark from rendered images of the protected 3DGS model. However, an unauthorized user may attempt to weaken or remove the watermark by applying distortions either to the rendered images or directly to the 3DGS model itself. Tab.~\ref{tab:wm-robustness} reports the watermark recovery results under these image- and model-level distortion scenarios. The results show that the proposed method preserves competitive watermark recovery performance even after such distortions, despite jointly optimizing watermark and adversarial signals.

\textbf{Robustness of adversarial noise.} In the application scenario of Fig.~\ref{fig:teaser}, Bob may attempt unauthorized editing using the acquired 3DGS model. Furthermore, if Bob is aware that adversarial optimization has been applied to deter such editing, he may try to weaken the protection by adding random noise to the Gaussians that compose the scene, or by performing operations such as random Gaussian cloning or random Gaussian pruning. Tab.~\ref{tab:adv-robustness} reports the results under these model-distortion scenarios and evaluates how successfully the proposed method continues to deter editing. The results suggest that the proposed framework preserves its edit-deterrence capability even under such distortions.
\begin{table*}[ht]
    \centering
    \caption{\textbf{Robustness of watermark recovery.} Bit acc. is measured in three settings: no distortion, image-level distortions on rendered images, and model-level distortions on the protected 3DGS.}
    \label{tab:wm-robustness}
    \small
    \setlength{\tabcolsep}{5pt}
    \renewcommand{\arraystretch}{1.25}
    \resizebox{\textwidth}{!}{%
    \begin{tabular}{lcccccccccc}
        \toprule
        \multirow{3}{*}{Method} & \multicolumn{10}{c}{Bit Acc. (\%) $\uparrow$} \\
        \cmidrule(lr){2-11}
         & \multirow{2}{*}{none} & \multicolumn{3}{c}{Model Distortion} & \multicolumn{6}{c}{Image Distortion} \\
        \cmidrule(lr){3-5} \cmidrule(lr){6-11}
         &  & noise & prune & clone & noise ($\sigma$=0.01) & rotation ($\pm \pi/6$) & scaling (75\%) & blur ($\sigma$=0.1) & crop (40\%) & jpeg ($Q$=50) \\
        \midrule
        3DGSW~\citep{3dgsw}          & \textbf{99.00} & \textbf{97.46} & \textbf{98.69} & \textbf{99.00} & \textbf{98.92} & \textbf{87.02} & \textbf{98.84} & \textbf{99.00} & \underline{85.36} & \textbf{94.82} \\
        GaussianMarker~\citep{gaussianmarker} & 98.51 & 78.27 & 92.58 & 95.24 & \underline{97.51} & 81.16 & 96.42 & 98.51 & 72.35 & 73.58 \\
        GuardSplat~\citep{guardsplt}     & \underline{98.92} & 64.49 & 80.16 & 89.26 & 96.73 & 81.61 & 94.95 & \underline{98.92} & \textbf{98.58} & 67.95 \\
        \rowcolor{gray!15}
        GuardMarkGS (Ours)           & 97.23 & \underline{92.94} & \underline{94.21} & \underline{97.23} & 96.73 & \underline{82.62} & \underline{96.97} & 97.23 & 71.05 & \underline{87.86} \\
        \bottomrule
    \end{tabular}
    }
\end{table*}
\begin{table*}[ht]
    \centering
    \caption{\textbf{Robustness of edit deterrence.} We apply model-level distortions to each protected 3DGS representation, perform 3D editing, and report CLIP, CLIP-T, and CLIP-D measured on the edited outputs. For each metric group, Orig. denotes the score measured from edits of the pretrained 3DGS, while Method reports scores measured from edits of the protected scene produced by each method.}
    \label{tab:adv-robustness}
    \small
    \setlength{\tabcolsep}{5pt}
    \renewcommand{\arraystretch}{1.25}
    \resizebox{\textwidth}{!}{%
    \begin{tabular}{lccccccccccccccc}
        \toprule
        \multirow{3}{*}{Method}
        & \multicolumn{5}{c}{CLIP $\downarrow$}
        & \multicolumn{5}{c}{CLIP-T $\downarrow$}
        & \multicolumn{5}{c}{CLIP-D $\downarrow$} \\
        \cmidrule(lr){2-6} \cmidrule(lr){7-11} \cmidrule(lr){12-16}
        & \multirow{2}{*}{Orig.} & \multicolumn{4}{c}{Method}
        & \multirow{2}{*}{Orig.} & \multicolumn{4}{c}{Method}
        & \multirow{2}{*}{Orig.} & \multicolumn{4}{c}{Method} \\
        \cmidrule(lr){3-6} \cmidrule(lr){8-11} \cmidrule(lr){13-16}
        &  & none & noise & prune & clone
        &  & none & noise & prune & clone
        &  & none & noise & prune & clone \\
        \midrule
        3DGSW~\citep{3dgsw}          
        & \multirow{5}{*}{1.0000} & 0.9629 & 0.9322 & 0.9545 & 0.9602 
        & \multirow{5}{*}{0.2825} & 0.2826 & 0.2761 & 0.2808 & 0.2822 
        & \multirow{5}{*}{0.1476} & 0.1408 & 0.1241 & 0.1330 & 0.1409 \\
        
        GaussianMarker~\citep{gaussianmarker} 
        &  & 0.9731 & 0.9443 & 0.9665 & 0.9710 
        &  & 0.2835 & 0.2774 & 0.2820 & 0.2838 
        &  & 0.1537 & 0.1396 & 0.1491 & 0.1554 \\
        
        GuardSplat~\citep{guardsplt}     
        &  & 0.9854 & 0.9521 & 0.9757 & 0.9822 
        &  & 0.2819 & 0.2761 & 0.2815 & 0.2830 
        &  & 0.1492 & 0.1281 & 0.1452 & 0.1499 \\
        
        DEGauss~\citep{degauss}        
        &  & \underline{0.9250} & \underline{0.8965} & \underline{0.9170} & \underline{0.9200} 
        &  & \underline{0.2713} & \underline{0.2654} & \underline{0.2697} & \underline{0.2706} 
        &  & \underline{0.1268} & \underline{0.1136} & \underline{0.1233} & \underline{0.1221} \\
        
        \rowcolor{gray!15}
        GuardMarkGS (Ours)           
        &  & \textbf{0.9093} & \textbf{0.8895} & \textbf{0.9081} & \textbf{0.9063} 
        &  & \textbf{0.2676} & \textbf{0.2631} & \textbf{0.2659} & \textbf{0.2677} 
        &  & \textbf{0.1086} & \textbf{0.1035} & \textbf{0.1086} & \textbf{0.1092} \\
        \bottomrule
    \end{tabular}
    }
\end{table*}

\section{Definition of CLIP-based edit-deterrence metrics}
\label{app:clip_metrics}
We use three CLIP-based metrics to quantify edit deterrence: CLIP, CLIP-T, and CLIP-D. All metrics are evaluated on the same novel-view set after applying the editing pipeline. In Tab.~\ref{tab:main} of the main paper, the \textit{Orig.} columns denote scores measured from edits of the original pretrained 3DGS scene, the \textit{Method} columns denote scores measured from edits of the protected scene produced by each method, and \textit{Diff.} denotes the gap between them.

Let $I_v^{\mathrm{edit}}$ be the edited render of the original scene at view $v$, and let $I_{i,v}^{\mathrm{edit}}$ be the edited render of the protected scene produced by method $i$. We use CLIP image and text encoders, denoted by $f_I(\cdot)$ and $f_T(\cdot)$, with cosine similarity between normalized embeddings.

\textbf{CLIP} measures the visual similarity between the edited protected result and the edited original result, with $o^{\mathrm{clip}}=1$ because the edited original result is compared with itself:
\begin{equation}
    r_i^{\mathrm{clip}}
    =
    \frac{1}{|\mathcal{V}_{\mathrm{eval}}|}
    \sum_{v\in\mathcal{V}_{\mathrm{eval}}}
    \cos\!\left(
    f_I(I_{i,v}^{\mathrm{edit}}),
    f_I(I_v^{\mathrm{edit}})
    \right).
\end{equation}

\textbf{CLIP-T} measures how well the edited result matches the target edit prompt $\psi_{\mathrm{tgt}}$:
\begin{equation}
    o^{\mathrm{clipT}}
    =
    \frac{1}{|\mathcal{V}_{\mathrm{eval}}|}
    \sum_{v}
    \cos\!\left(f_I(I_v^{\mathrm{edit}}), f_T(\psi_{\mathrm{tgt}})\right),
    \qquad
    r_i^{\mathrm{clipT}}
    =
    \frac{1}{|\mathcal{V}_{\mathrm{eval}}|}
    \sum_{v}
    \cos\!\left(f_I(I_{i,v}^{\mathrm{edit}}), f_T(\psi_{\mathrm{tgt}})\right).
\end{equation}

\textbf{CLIP-D} measures whether the image-editing direction aligns with the text-editing direction. Let $I_v^{\mathrm{src}}$ and $I_{i,v}^{\mathrm{src}}$ denote the corresponding pre-edit renders, and let $\psi_{\mathrm{src}}$ and $\psi_{\mathrm{tgt}}$ be the source and target edit prompts, respectively. We define the text and image directions in CLIP space as
\begin{equation}
    d_t
    =
    \operatorname{norm}\!\left(f_T(\psi_{\mathrm{tgt}})-f_T(\psi_{\mathrm{src}})\right),
    \qquad
    d_v^{\mathrm{img}}
    =
    \operatorname{norm}\!\left(f_I(I_v^{\mathrm{edit}})-f_I(I_v^{\mathrm{src}})\right),
\end{equation}
\begin{equation}
    d_{i,v}^{\mathrm{img}}
    =
    \operatorname{norm}\!\left(f_I(I_{i,v}^{\mathrm{edit}})-f_I(I_{i,v}^{\mathrm{src}})\right),
\end{equation}
where $\operatorname{norm}(\cdot)$ denotes $\ell_2$ normalization. The CLIP-D scores are then
\begin{equation}
    o^{\mathrm{clipD}}
    =
    \frac{1}{|\mathcal{V}_{\mathrm{eval}}|}
    \sum_{v}
    \cos(d_v^{\mathrm{img}}, d_t),
    \qquad
    r_i^{\mathrm{clipD}}
    =
    \frac{1}{|\mathcal{V}_{\mathrm{eval}}|}
    \sum_{v}
    \cos(d_{i,v}^{\mathrm{img}}, d_t).
\end{equation}

For each metric $k\in\{\mathrm{clip},\mathrm{clipT},\mathrm{clipD}\}$, we report the difference
\begin{equation}
    \Delta_i^k = o^k - r_i^k.
\end{equation}
A lower method score $r_i^k$ indicates that the edited protected result is less aligned with the intended edit. Accordingly, a larger positive difference $\Delta_i^k$ indicates a larger reduction from the edited original result, and therefore stronger edit deterrence.

\section{Definition of sUCPS}
\label{app:sucps}
To evaluate the effectiveness of copyright-protection methods, we define the \emph{soft Unified Copyright Protection Score} (sUCPS). This metric jointly reflects practical 3D copyright-protection requirements: ownership traceability, edit deterrence, and rendering fidelity. Rather than evaluating these aspects in isolation, sUCPS rewards methods that maintain balanced performance across all three. The following definitions specify how these aspects are normalized and aggregated into sUCPS.

Let $i$ index a protection method. We define three normalized subscores: a traceability score $T_i$, an edit-deterrence score $E_i$, and a fidelity score $F_i$. All subscores are mapped to $[0,1]$, where larger values indicate better performance. For compactness, we use $\operatorname{GM}(\cdot)$ and $\operatorname{HM}(\cdot)$ to denote the geometric and harmonic means, respectively, and clip all normalized values to $[0,1]$.

\textbf{Traceability score.} Let $b_i \in [0,1]$ be the watermark bit accuracy of method $i$. For example, a bit accuracy of $97.23\%$ is used as $b_i=0.9723$ in the sUCPS computation. For methods that do not provide watermark decoding, we assign the chance-level reference $b_i=0.5$. Let $b_{\max}$ be the best bit accuracy among methods that support watermark decoding. We define
\begin{equation}
    T_i
    =
    \operatorname{clip}_{[0,1]}
    \left[
    \frac{1}{2}
    \left(
    1+
    \frac{b_i-0.5}{b_{\max}-0.5}
    \right)
    \right].
\end{equation}
This maps chance-level ownership tracing to a neutral reference and assigns higher scores as watermark recovery improves, with bit accuracy approaching the best observed value.

\textbf{Edit-deterrence score.} For each CLIP-based edit-deterrence metric, we measure the gap between the edited original scene and the edited protected scene. Let
\begin{equation}
    \Delta_i^{\mathrm{clip}} = o^{\mathrm{clip}} - r_i^{\mathrm{clip}},
    \qquad
    \Delta_i^{\mathrm{clipT}} = o^{\mathrm{clipT}} - r_i^{\mathrm{clipT}},
    \qquad
    \Delta_i^{\mathrm{clipD}} = o^{\mathrm{clipD}} - r_i^{\mathrm{clipD}},
\end{equation}
where $o$ denotes the score of the edited original scene and $r_i$ the score of the edited protected result from method $i$. A larger positive gap indicates stronger suppression of the editing outcome. We normalize each gap, while preserving its sign, by the largest absolute gap among compared methods:
\begin{equation}
    u_i^{k}
    =
    \operatorname{clip}_{[0,1]}
    \left[
    \frac{1}{2}
    \left(
    1+
    \frac{\Delta_i^k}{\max_j |\Delta_j^k|}
    \right)
    \right],
    \qquad
    k\in\{\mathrm{clip},\mathrm{clipT},\mathrm{clipD}\}.
\end{equation}
The edit-deterrence score is then defined as
\begin{equation}
    E_i
    =
    \operatorname{GM}
    \left(
    u_i^{\mathrm{clip}},
    u_i^{\mathrm{clipT}},
    u_i^{\mathrm{clipD}}
    \right).
\end{equation}

\textbf{Fidelity score.} For rendering fidelity, we use PSNR, SSIM, and LPIPS measured on protected novel-view rendered images. Since higher PSNR and SSIM are better while lower LPIPS is better, we define the normalized fidelity scores as
\begin{equation}
    p_i=\frac{\mathrm{PSNR}_i}{\max_j \mathrm{PSNR}_j},
    \qquad
    q_i=\frac{\mathrm{SSIM}_i}{\max_j \mathrm{SSIM}_j},
    \qquad
    \ell_i=\frac{\min_j \mathrm{LPIPS}_j}{\mathrm{LPIPS}_i}.
\end{equation}
We aggregate them using a geometric mean:
\begin{equation}
    F_i
    =
    \operatorname{GM}(p_i,q_i,\ell_i).
\end{equation}

Finally, we combine the three subscores using a harmonic mean:
\begin{equation}
    \mathrm{sUCPS}_i
    =
    \operatorname{HM}(T_i,E_i,F_i).
    \label{eq:sucps}
\end{equation}
The harmonic mean penalizes methods that perform well in only one or two aspects but fail in the remaining one. Thus, a high sUCPS indicates that a method jointly preserves ownership traceability, edit deterrence, and rendering fidelity.

\section{Cross-attention diversion}
\label{app:xattn}
In addition to latent-anchor separation and denoising-trajectory diversion introduced in Sec.~\ref{subsec:adversarialobjective} of the main paper, our adversarial branch includes a cross-attention diversion term. It aims to disturb how the diffusion editor aligns the edit prompt with the rendered input, while keeping the editor fixed.
 
For a sampled view $v$, we first render the protected scene and the frozen reference scene,
\begin{equation}
    I_v^{\mathrm{prot}}=\mathcal{R}_v(\mathcal{G}^{\mathrm{prot}}),
    \qquad
    I_v^{\mathrm{ref}}=\mathcal{R}_v(\mathcal{G}^{\mathrm{ref}}),
\end{equation}
and encode them into the latent space of the diffusion editor,
\begin{equation}
    z_v^{\mathrm{prot}} = z(I_v^{\mathrm{prot}}),
    \qquad
    z_v^{\mathrm{ref}} = z(I_v^{\mathrm{ref}}).
\end{equation}
We then construct noisy latents using the same edit prompt $\psi$, timestep $t$, and sampled noise $\epsilon$:
\begin{equation}
    \tilde z_v^{\mathrm{prot}} = \alpha_t z_v^{\mathrm{prot}} + \sigma_t \epsilon,
    \qquad
    \tilde z_v^{\mathrm{ref}} = \alpha_t z_v^{\mathrm{ref}} + \sigma_t \epsilon.
\end{equation}
Here, $\alpha_t$ and $\sigma_t$ are the standard diffusion noise-schedule coefficients at timestep $t$. Sharing $(\psi,t,\epsilon)$ ensures that the resulting diversion signal is driven by the scene difference itself, rather than by stochastic variation inside the editor.

We compute the cross-attention descriptor from the U-Net mid block. Lying at the denoising U-Net's bottleneck, the mid block features have a broader receptive field and encode more semantic, prompt-dependent structure than early high-resolution layers~\citep{facesheild}. Perturbing the mid-block cross-attention disrupts semantic correspondence between the edit prompt and rendered image, helping divert the intended editing trajectory while reducing reliance on superficial texture-level perturbations. Let
\begin{equation}
    Q_{\mathrm{mid},v}^{\mathrm{prot}} = W_{\mathrm{mid}}^{Q}\!\big(H_{\mathrm{mid}}(\tilde z_v^{\mathrm{prot}},\psi,t)\big),
    \qquad
    Q_{\mathrm{mid},v}^{\mathrm{ref}} = W_{\mathrm{mid}}^{Q}\!\big(H_{\mathrm{mid}}(\tilde z_v^{\mathrm{ref}},\psi,t)\big),
\end{equation}
where $H_{\mathrm{mid}}(\cdot)$ denotes the hidden representation entering the mid-block text-conditioned cross-attention and $W_{\mathrm{mid}}^{Q}$ is its query projection. To summarize this response for each rendered view, we average the $N_q$ query tokens, where $Q_{n,:}$ denotes the feature vector of the $n$-th token:
\begin{equation}
    q_{\mathrm{mid},v}^{\mathrm{prot}} = \rho\!\left(Q_{\mathrm{mid},v}^{\mathrm{prot}}\right),
    \qquad
    q_{\mathrm{mid},v}^{\mathrm{ref}} = \rho\!\left(Q_{\mathrm{mid},v}^{\mathrm{ref}}\right),
    \qquad
    \rho(Q)=\frac{1}{N_q}\sum_{n=1}^{N_q}Q_{n,:}.
\end{equation}

We define the cross-attention diversion term for view $v$ as
\begin{equation}
    \mathcal{S}_{\mathrm{xattn}}^{(v)}
    =
    \frac{1}{d_{\mathrm{mid}}}
    \left\|
    q_{\mathrm{mid},v}^{\mathrm{prot}}
    -
    q_{\mathrm{mid},v}^{\mathrm{ref}}
    \right\|_2^2,
    \label{eq:xattn}
\end{equation}
where $d_{\mathrm{mid}}$ is the query dimension of the mid-block descriptor. Since $\mathcal{S}_{\mathrm{xattn}}^{(v)}$ is subtracted in Eq.~\eqref{eq:adv}, maximizing this diversion term encourages the protected render to induce query representations that differ from those of the frozen reference scene under the same prompt and diffusion condition. This perturbs the prompt-to-image alignment used by the editor and helps divert the editing trajectory.

Importantly, both the diffusion editor and the reference scene remain frozen throughout optimization. Thus, the gradient of $\mathcal{S}_{\mathrm{xattn}}^{(v)}$ flows only through the rendered protected image and ultimately updates the selected Gaussian parameters of $\mathcal{G}^{\mathrm{prot}}$.

\begin{figure*}[t]
    \centering
    \begin{subfigure}[t]{0.30\linewidth}
        \centering
        \includegraphics[width=0.75\linewidth]{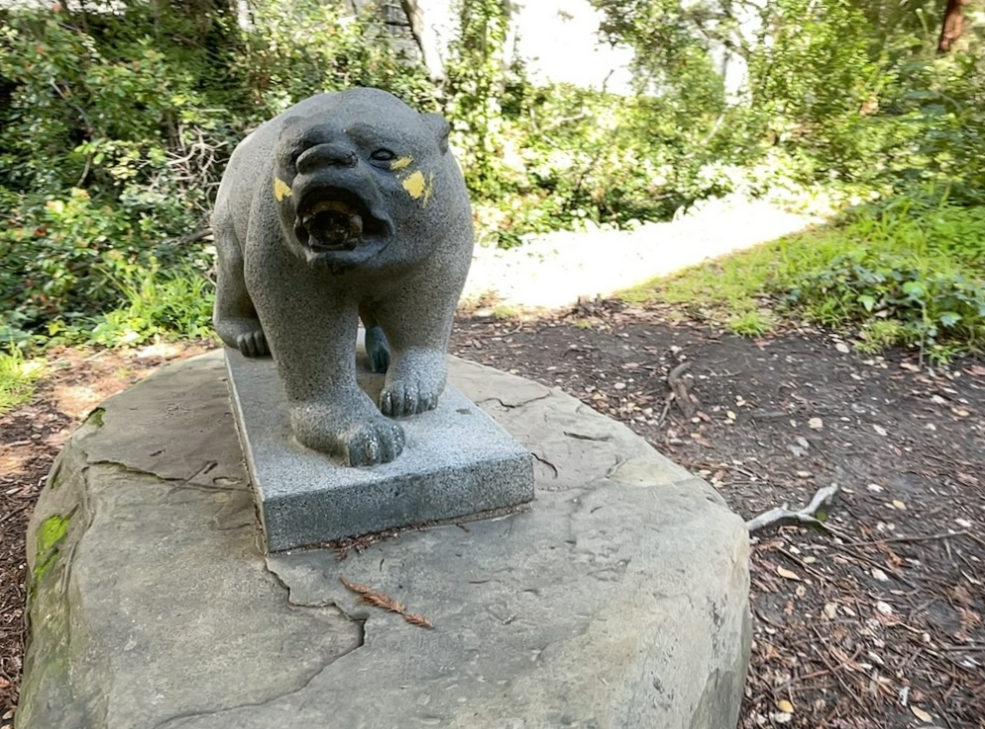}
        \caption{Same-view image.}
        \label{fig:xattn_vis_render}
    \end{subfigure}
    \hfill
    \begin{subfigure}[t]{0.67\linewidth}
        \centering
        \includegraphics[width=\linewidth]{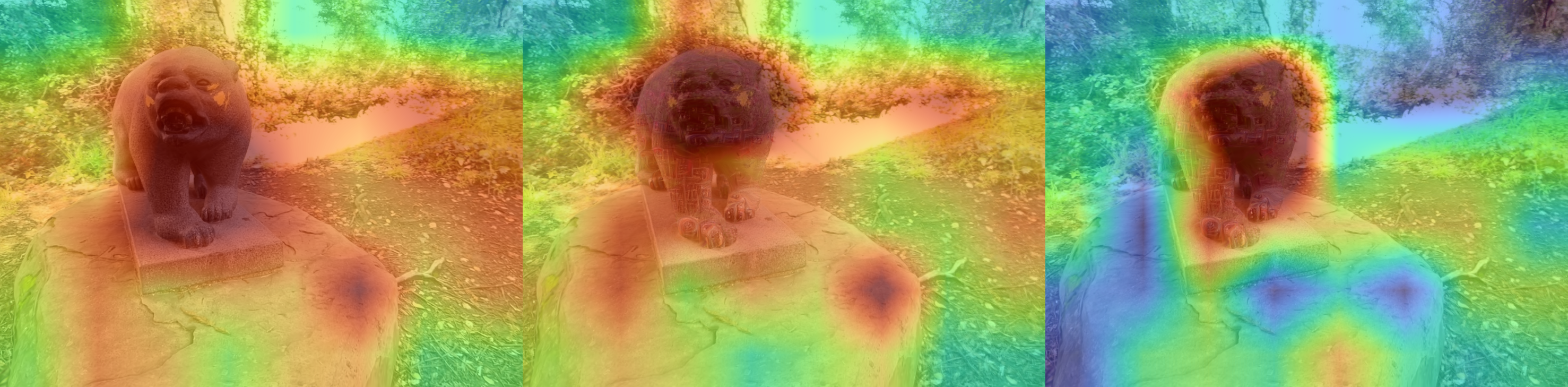}
        \caption{Reference, protected, and difference maps.}
        \label{fig:xattn_vis_maps}
    \end{subfigure}
    \caption{\textbf{Visualization of cross-attention diversion.} Left: the image from the same viewpoint shown without an attention overlay. Right: visualization of the cross-attention map for \textit{statue}. The three panels overlay the token attention on the frozen reference render, the protected render, and their absolute difference, respectively. The difference panel shows a large attention change around \textit{statue}, indicating that the protected scene changes the editor's prompt-conditioned spatial focus.}
    \label{fig:xattn_vis}
\vspace{-15pt}
\end{figure*}

Fig.~\ref{fig:xattn_vis} provides a visualization of the effect of cross-attention diversion. Fig.~\ref{fig:xattn_vis_render} shows the image from the same viewpoint without an attention overlay. Fig.~\ref{fig:xattn_vis_maps} shows the cross-attention map for the prompt \textit{statue}. The third panel in Fig.~\ref{fig:xattn_vis_maps} visualizes the difference between the two maps, where red regions indicate areas in which the 3D editing model exhibits larger changes in attention to the given prompt compared with the original image. This reveals a strong change around \textit{statue}, suggesting that the prompt-to-image alignment pathway used by the diffusion editor has been meaningfully altered.

\section{Additional ablation on masking strategy}
\label{app:ablation}
In this section, we provide additional ablation results on the Gaussian masking strategy introduced in Sec.~\ref{subsec:update-saliency-selection} of the main paper. Instruction-driven 3D editing can target not only salient foreground objects, such as changing the identity, color, or appearance of a main subject, but also background regions, such as altering the scene style, atmosphere, or surrounding environment. In the main paper, we describe an update-saliency-motivated Gaussian selection strategy that identifies a mask-selected Gaussian subset and injects stronger adversarial signals into it. However, the mask is implemented in a soft manner rather than as a strict hard mask, allowing Gaussians that fall below the hard selection threshold but still have nonzero mask overlap to receive weaker adversarial updates.

\begin{figure*}[t]
    \centering
    \begin{minipage}{\textwidth}
        \centering
        \includegraphics[width=\linewidth]{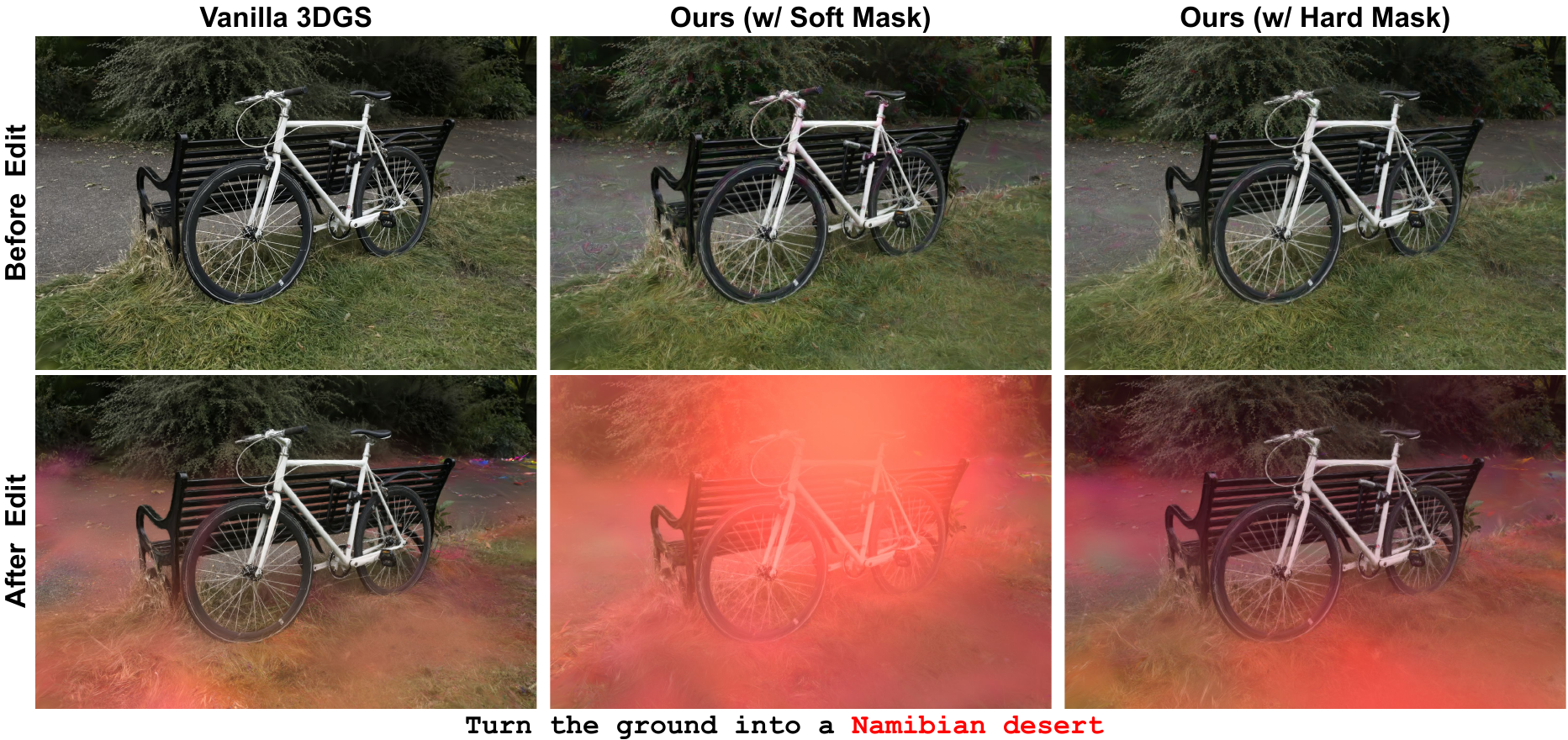}
        \caption{\textbf{Ablation on hard and soft Gaussian masking.} Hard masking confines adversarial updates to the selected region, whereas soft mask concentrates updates on selected Gaussians while retaining adversarial influence on surrounding Gaussians. This design is more suitable for editing prompts that affect not only foreground objects but also background regions, scene style, or atmosphere.}
        \label{fig:soft_mask}
    \end{minipage}
\vspace{-15pt}
\end{figure*}

Fig.~\ref{fig:soft_mask} compares the defense performance of hard and soft masking under edits that affect background regions or the surrounding scene context. The comparison shows that, in this setting, soft masking is more effective than hard masking because it avoids strictly confining adversarial perturbations to the selected region. Together with the selection ablation in Tab.~\ref{tab:joint-design-ablation} of the main paper, this suggests that selecting a Gaussian subset for adversarial noise injection helps balance edit deterrence, watermark recovery, and rendering fidelity, and is better implemented in a soft rather than hard manner.

\section{Additional ablation on adversarial parameter groups}
\label{app:param_group}
While the update-saliency-motivated Gaussian selection strategy determines \emph{which Gaussians} should receive adversarial signals, we additionally examine \emph{which parameters} of the selected Gaussians should be perturbed to achieve effective edit deterrence. Following prior 3DGS watermark fine-tuning practice, we keep the mean-position $\mu_i$ fixed, since directly changing Gaussian centers can easily affect scene geometry and rendering quality~\citep{3dgsw}. We therefore focus this ablation on the remaining geometry parameters, scale $s_i$ and rotation $r_i$, and the appearance parameters, opacity $\alpha_i$ and color $\mathbf{c}_i$. Tab.~\ref{tab:param-group} reports the resulting parameter-group ablation and shows that effective edit deterrence is achieved when adversarial noise is injected into all of these non-position parameter groups.

\section{Watermark recovery after 3D editing}
\label{app:wm_after_dge}
We further evaluate the watermark recovery performance of existing 3DGS-based watermarking methods after instruction-driven 3D editing~\citep{dge}. For each watermarking baseline, we first protect a pretrained 3DGS scene using the corresponding method, perform editing with DGE~\citep{dge}, and then render the edited scene from the same novel-view evaluation set used in the main experiments. We decode the watermark from these rendered images and report the resulting bit accuracy in Tab.~\ref{tab:wm-after-dge}.
\begin{figure*}[ht]
    \centering
    \begin{minipage}[t]{0.485\textwidth}
        \captionsetup{type=table}
        \caption{\textbf{Ablation on adversarial noise injection across Gaussian parameter groups.} We ablate the geometry and appearance parameter groups used for noise injection. The results suggest that injecting adversarial noise into scale, rotation, opacity, and color achieves the strongest edit deterrence while maintaining competitive watermark detectability and rendering quality.}
        \label{tab:param-group}
        \centering
        \footnotesize
        \begingroup
        \setlength{\tabcolsep}{5pt}
        \renewcommand{\arraystretch}{1.0}
        \setlength{\aboverulesep}{0.2ex}
        \setlength{\belowrulesep}{0.2ex}
        \setlength{\cmidrulesep}{0.1ex}
        \resizebox{\linewidth}{!}{%
        \begin{tabular}{ccccccc}
            \toprule
            \multicolumn{2}{c}{Geometry} & \multicolumn{2}{c}{Appearance} & Bit Acc. (\%) $\uparrow$ & CLIP $\downarrow$ & PSNR $\uparrow$ \\
            \cmidrule(lr){1-2} \cmidrule(lr){3-4}
            $s_i$ & $r_i$ & $\alpha_i$ & $\mathbf{c}_i$ &  &  &  \\
            \midrule
            $\checkmark$ & $\checkmark$ & - & - & 97.51 & 0.9658 & 31.60 \\
            - & - & $\checkmark$ & - & 97.39 & 0.9734 & \underline{31.81} \\
            - & - & - & $\checkmark$ & \underline{97.54} & 0.9631 & \textbf{32.03} \\
            - & - & $\checkmark$ & $\checkmark$ & 97.21 & 0.9486 & 31.18 \\
            $\checkmark$ & $\checkmark$ & - & $\checkmark$ & 97.04 & \underline{0.9322} & 30.86 \\
            $\checkmark$ & $\checkmark$ & $\checkmark$ & - & \textbf{97.63} & 0.9650 & 31.03 \\
            \rowcolor{gray!15}
            $\checkmark$ & $\checkmark$ & $\checkmark$ & $\checkmark$ & 97.23 & \textbf{0.9093} & 30.36 \\
            \bottomrule
        \end{tabular}
        }
        \endgroup
    \end{minipage}
    \hfill
    \begin{minipage}[t]{0.485\textwidth}
        \captionsetup{type=table}
        \caption{\textbf{Bit accuracy after 3D editing.} We report watermark bit accuracy before and after applying DGE, an instruction-driven 3D editing method. The Drop (pp) is computed as the decrease from the pre-edit bit accuracy to the post-edit bit accuracy, measured in percentage points.}
        \label{tab:wm-after-dge}
        \vspace{15pt}
        \centering
        \footnotesize
        \begingroup
        \setlength{\tabcolsep}{5pt}
        \renewcommand{\arraystretch}{1.75}
        \setlength{\aboverulesep}{0.2ex}
        \setlength{\belowrulesep}{0.2ex}
        \setlength{\cmidrulesep}{0.1ex}
        {\raggedleft\scriptsize $\mathrm{Drop}=\mathrm{Before\ edit}-\mathrm{After\ edit}$\par}
        \resizebox{\linewidth}{!}{%
        \begin{tabular}{lccc}
            \toprule
            \multirow{2}{*}{Method} & \multicolumn{2}{c}{Bit Acc. (\%) $\uparrow$} & \multirow{2}{*}{Drop (pp) $\downarrow$} \\
            \cmidrule(lr){2-3}
             & Before edit & After edit &  \\
            \midrule
            3DGSW~\citep{3dgsw} & \textbf{99.00} & \textbf{89.32} & \cellcolor{orange!10}\textbf{9.68} \\
            GaussianMarker~\citep{gaussianmarker} & 98.51 & \underline{75.02} & \cellcolor{orange!25}\underline{23.49} \\
            GuardSplat~\citep{guardsplt} & \underline{98.92} & 54.58 & \cellcolor{red!20}44.34 \\
            \bottomrule
        \end{tabular}
        }
        \endgroup
    \end{minipage}
\vspace{-10pt}
\end{figure*}

As shown in Tab.~\ref{tab:wm-after-dge}, existing 3DGS-based watermarking methods exhibit non-negligible drops in bit accuracy after editing. Specifically, 3DGSW, GaussianMarker, and GuardSplat show decreases of 9.68 pp, 23.49 pp, and 44.34 pp, respectively, indicating that the DGE edit-update process can substantially degrade watermark recoverability.

However, the point of this experiment is not merely to measure how well a watermark can be recovered after editing. Even if a watermarking method were to maintain nearly unchanged bit accuracy after 3D editing, watermarking would still remain a passive form of protection that only provides post-hoc ownership evidence. Therefore, practical 3DGS copyright protection requires more than ownership tracing alone. This motivates a unified protection framework that jointly optimizes recoverable ownership tracing and active edit deterrence within a single protected 3DGS representation.

\section{Additional qualitative results}
\label{app:qualitative}
\begin{figure*}[t]
    \centering
    \begin{minipage}{\textwidth}
        \centering
        \includegraphics[width=\linewidth]{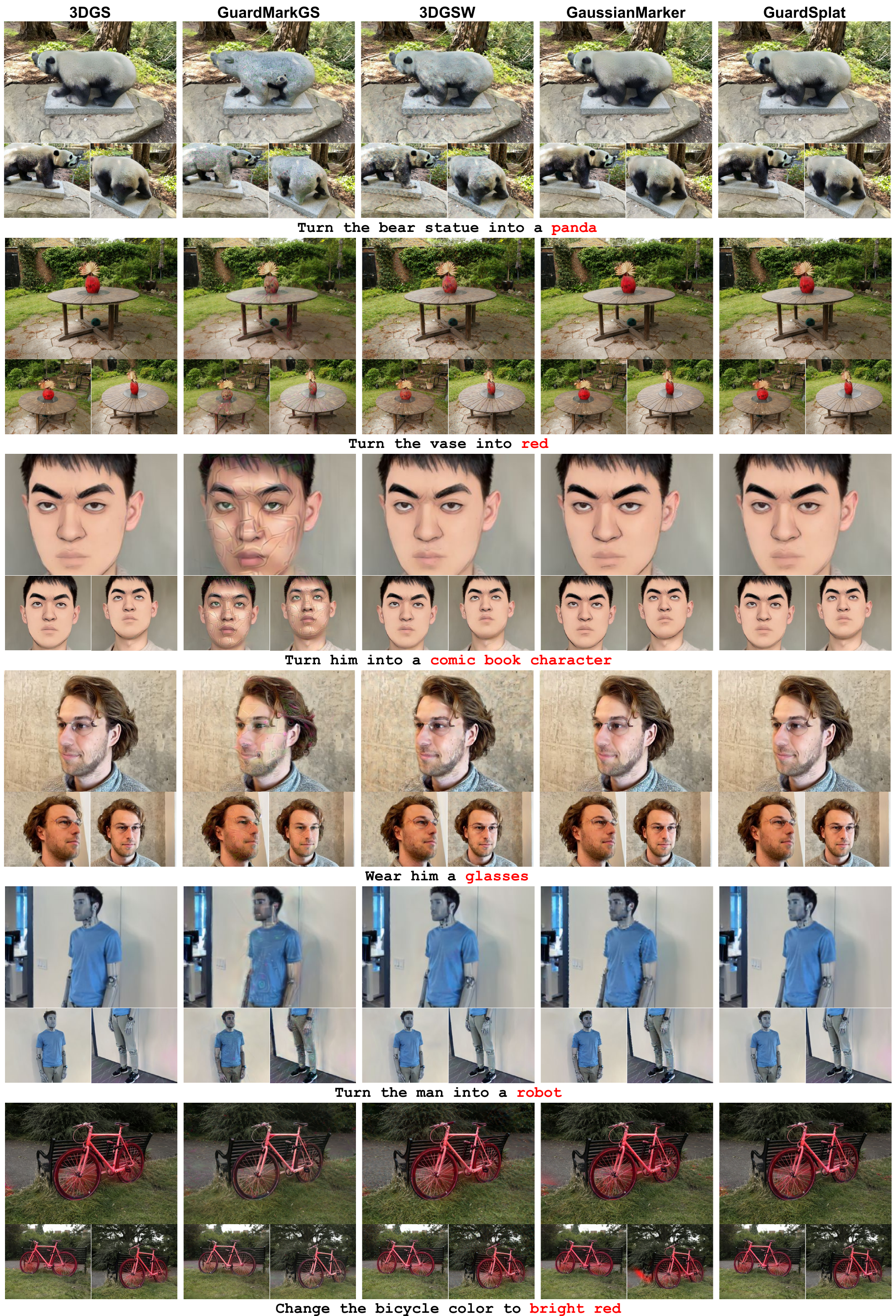}
        \caption{\textbf{Additional qualitative results.} We show the editing results for each scene using DGE~\citep{dge}.}
        \label{fig:additional_qualitative}
    \end{minipage}
\vspace{-15pt}
\end{figure*}

Fig.~\ref{fig:additional_qualitative} shows additional qualitative results beyond those shown in the main paper.



\end{document}